%% file: main.tex
\documentclass[manuscript,nonacm]{acmart}
\usepackage{xcolor}
\usepackage{colortbl}
\definecolor{mygray}{HTML}{f0f0f0}
\usepackage{tabulary,multirow,overpic,xcolor}
\usepackage{graphicx}
\usepackage{booktabs}
\usepackage{threeparttable}
\usepackage{bbding}
\usepackage{makecell}
\usepackage{subcaption}
\usepackage{amsmath}
\AtBeginDocument{%
  }

\begin{document}

\title{An Egocentric Vision-Language Model based Portable Real-time Smart Assistant}

% \author{Yifei Huang$^1$, Jilan Xu$^2$, Baoqi Pei$^3$, Yuping He$^4$, Guo Chen$^4$, Lijin Yang$^1$, Xinyuan Chen, Yaohui Wang, Zheng Nie, Jinyao Liu, Guoshun Fan, Dechen Lin, Fang Fang, Kunpeng Li, Chang Yuan, Yali Wang, Yu Qiao, Limin Wang$^4$}
% \affiliation{%
%   \institution{Shanghai AI Laboratory, $^1$The University of Tokyo, $^2$Fudan University, $^3$Zhejiang University, $^4$Nanjing University}
%   \country{China}
%   }
% \email{hyf015@gmail.com}

\author{Yifei Huang}
\affiliation{%
  \institution{The University of Tokyo}
  \city{Tokyo}
  \country{Japan}
}
\email{hyf@iis.u-tokyo.ac.jp}
\author{Jilan Xu}
\affiliation{%
  \institution{Fudan University}
  \city{Shanghai}
  \country{China}
}
\author{Baoqi Pei}
\affiliation{%
  \institution{Zhejiang University}
  \city{Hangzhou}
  \country{China}
}

\author{Yuping He}
\author{Guo Chen}
\affiliation{%
  \institution{Nanjing University}
  \city{Nanjing}
  \country{China}
}

\author{Mingfang Zhang}
\author{Lijin Yang}
\affiliation{%
  \institution{The University of Tokyo}
  \city{Tokyo}
  \country{Japan}
}

\author{Zheng Nie}
\author{Jinyao Liu}
\author{Guoshun Fan}
\author{Dechen Lin}
\author{Fang Fang}
\author{Kunpeng Li}
\author{Chang Yuan}
\author{Yaohui Wang}
\author{Xinyuan Chen}
\author{Yali Wang}
\author{Yu Qiao}
\author{Limin Wang}
\affiliation{%
  \institution{Shanghai AI Laboratory}
  \city{Shanghai}
  \country{China}
}

% \affiliation{%

% \affiliation{%

% \affiliation{%

% \affiliation{%

% \affiliation{%

\renewcommand{\shortauthors}{Huang et al.}
% \shortauthors

\begin{abstract}
We present Vinci, a vision-language system designed to provide real-time, comprehensive AI assistance on portable devices. At its core, Vinci leverages EgoVideo-VL, a novel model that integrates an egocentric vision foundation model with a large language model (LLM), enabling advanced functionalities such as scene understanding, temporal grounding, video summarization, and future planning. To enhance its utility, Vinci incorporates a memory module for processing long video streams in real time while retaining contextual history, a generation module for producing visual action demonstrations, and a retrieval module that bridges egocentric and third-person perspectives to provide relevant how-to videos for skill acquisition. Unlike existing systems that often depend on specialized hardware, Vinci is hardware-agnostic, supporting deployment across a wide range of devices, including smartphones and wearable cameras. In our experiments, we first demonstrate the superior performance of EgoVideo-VL on multiple public benchmarks, showcasing its vision-language reasoning and contextual understanding capabilities. We then conduct a series of user studies to evaluate the real-world effectiveness of Vinci, highlighting its adaptability and usability in diverse scenarios. We hope Vinci can establish a new framework for portable, real-time egocentric AI systems, empowering users with contextual and actionable insights. Including the frontend, backend, and models, all codes of Vinci are available at \url{https://github.com/OpenGVLab/vinci}.
\end{abstract}
% We demonstrate Vinci’s potential applications in , highlighting its ability to adapt across diverse scenarios without domain-specific tuning. 




\maketitle

\section{INTRODUCTION}
Wearable AI assistants are transforming human-computer interaction by offering real-time, context-aware support through seamless interaction~\cite{zhao2023unveiling,qi2023cases}. Advances in large language models (LLMs) have enabled natural language-based interfaces for various applications, including healthcare~\cite{yang2024drhouse}, navigation~\cite{chen2024enabling}, and social assistance~\cite{yang2024socialmind}. However, most existing AI assistants rely primarily on language as their central modality. In these systems, multimodal inputs such as images are typically converted into text, and responses are delivered in textual or speech-based formats This reliance on language limits the system's ability to fully exploit the rich semantic content of multimodal data. Moreover, converting non-linguistic data (\textit{e.g.}, images) into text often results in a significant loss of information, limiting these assistants' effectiveness in real-world applications.

\begin{figure}
    \centering
\includegraphics[width=\linewidth]{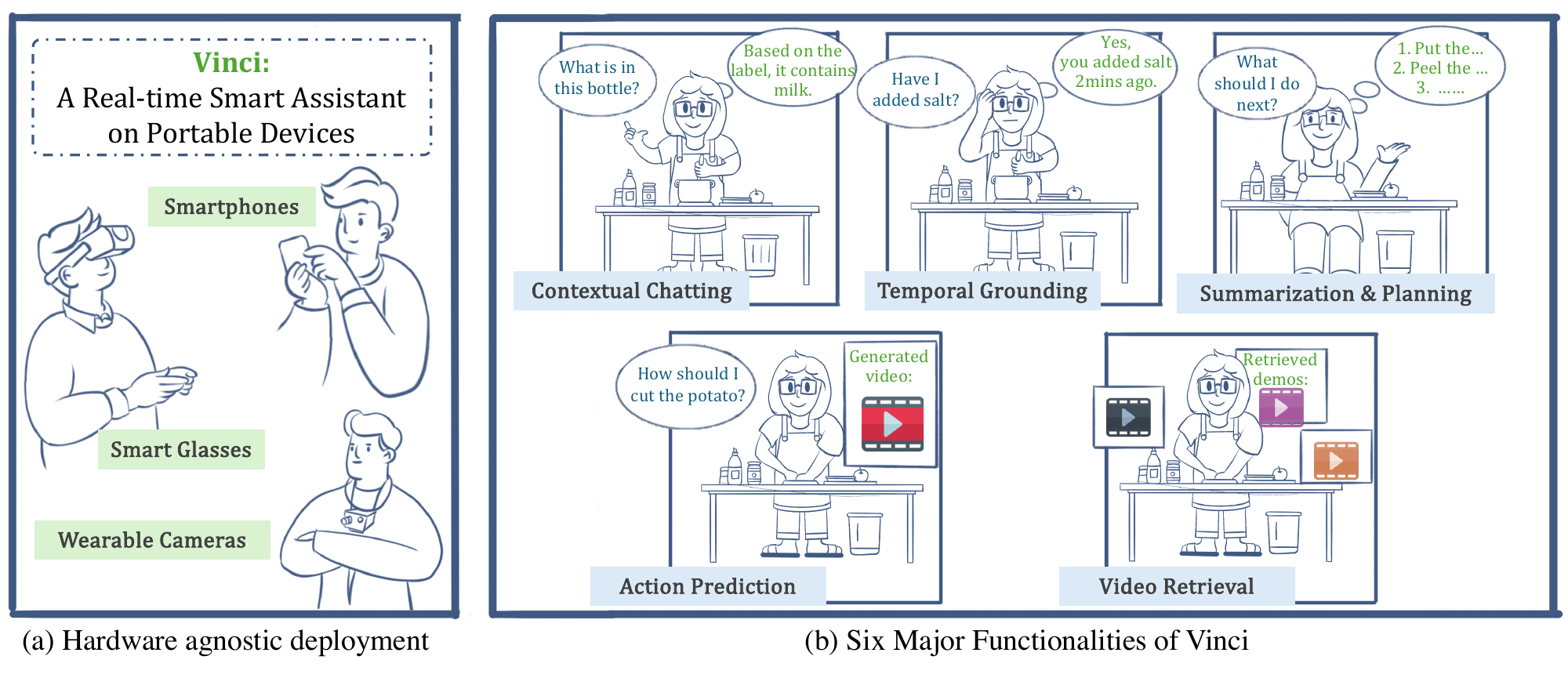}
    \caption{(a) \texttt{Vinci}'s hardware-agnostic deployment across various devices, including smart glasses, wearable cameras, and smartphones. (b) \texttt{Vinci}'s 6 core functionalities: contextual chatting for interactive queries, temporal grounding for reasoning over past events, summarization for concise activity overviews, planning for multi-step task execution, action prediction and video retrieval for providing visible skill demonstrations.}
    \label{fig:overview}
\end{figure}

While the recent advances in vision-language models (VLMs) show promise in combining visual and linguistic understanding~\cite{internvl,internvideo2,liu2023llava,zhang2024internlm,li2024llava}, directly deploying VLMs as wearable assistants presents several challenges. First, most VLMs focus primarily on analyzing external environments, but cannot infer unobservable states, such as the user’s intentions, past activities, or implicit goals. Second, most VLMs require high computational resources, making them impractical for real-time, continuous interactions in dynamic environments. Third, they lack memory mechanisms, which are essential for contextualizing and personalizing long-term user activities and enabling historical reasoning. Finally, they cannot generate visual guidance, which limits their effectiveness in providing intuitive, visually guided assistance. These challenges underscore the need for a comprehensive, real-time vision-language system specifically designed for wearable devices that can overcome these limitations and offer robust assistance in diverse scenarios.

To address these challenges, we introduce \texttt{Vinci}, a real-time, egocentric, multimodal vision-language assistant designed for portable and wearable devices.
We begin by conducting a user survey with 107 participants from diverse backgrounds, identifying key needs for wearable AI assistants.
Designed to meet these unique demands, \texttt{Vinci} tackles the aforementioned key limitations of existing approaches.
First, \texttt{Vinci} integrates a novel model, EgoVideo-VL, which combines the strengths of an egocentric vision foundation model with a large language model (LLM) to achieve robust multimodal understanding and interaction. 
Trained on egocentric data, this model is uniquely equipped to reason about both the external environment and the unobservable states and intentions of the system wearer.
Second, \texttt{Vinci} is designed for real-time performance, ensuring low latency for dynamic, continuous interactions. Third, \texttt{Vinci} integrates a memory module that retains and analyzes historical data, facilitating long-term temporal reasoning and enabling personalized interactions by adapting to the user's activity history. Lastly, \texttt{Vinci} extends its functionality with a generation module and a retrieval module, which provide visual demonstrations and actionable guidance in an intuitive and easily interpretable format, bridging the gap between understanding and user-friendly assistance.

With this carefully designed architecture, Vinci supports a broad range of capabilities tailored for real-time, context-aware interactions:
\begin{itemize}
    \item \textbf{Contextual Chatting.} Vinci enables natural, vision-grounded conversations, allowing users to ask questions about their surroundings and receive responses based on their current visual and temporal context.
    \item \textbf{Temporal Grounding.} Vinci’s memory module supports temporal grounding, empowering users to query past events, such as "When and where did I lose my keys?"
    \item \textbf{Summarization.} Vinci can analyze long video sequences, summarizing key moments into a concise, structured narrative. 
    \item \textbf{Future Planning.} Vinci assists users in planning future tasks by integrating predictive reasoning with insights from historical context, enabling step-by-step task planning. 
    \item \textbf{Action Prediction.} Vinci’s generation module produces realistic action demonstrations, helping users visually understand how to perform tasks. This approach ensures that guidance is not only actionable but also seamlessly aligned with the user’s current surroundings.
    \item \textbf{Video Retrieval.} Vinci’s retrieval module sources relevant third-person instructional videos, complementing egocentric guidance with expert demonstrations.
\end{itemize} 
Together, these functions establish Vinci as a versatile and powerful assistant, designed to enhance productivity, learning, and everyday decision-making.

% To evaluate Vinci’s performance across its diverse capabilities, we designed specific experiments to rigorously test each function. In addition to quantitative assessments, we conducted user studies by hiring participants to interact with the system in real-world scenarios. Participants were asked to perform tasks and provide feedback on their experience through a detailed questionnaire. The results were highly encouraging: 120\% of users expressed satisfaction with Vinci’s functionality, while 120\% reported that using the system can enhance their quality of life and work efficiency.
To rigorously evaluate Vinci’s real-world performance, we conduct both quantitative experiments and in-situ user studies. Participants interact with Vinci across diverse scenarios, and their feedback is collected through structured questionnaires. The results were highly encouraging: 90\% of users expressed satisfaction with Vinci’s functionality, while 85\% reported that using the system can enhance their quality of life and work efficiency.

In summary, the contributions of this paper are as follows:
\begin{itemize}
    \item We introduce Vinci, a real-time, hardware-agnostic vision-language assistant that operates on wearable devices and smartphones.
    \item We propose EgoVideo-VL, a user-centric vision-language model trained on egocentric data, enabling Vinci to understand both the external environment and the unobservable states of the wearer.
    \item We implement a suite of novel functionalities, including context-based chatting, temporal reasoning, multimodal summarization, future planning, visual action generation, and video retrieval for skill learning.
    \item We conduct rigorous evaluations through quantitative experiments and in-situ user studies, demonstrating Vinci’s effectiveness in real-world deployment scenarios.
\end{itemize}

\section{RELATED WORKS}
\subsection{LLMs and VLMs}

Large Vision-language models (VLMs) have achieved significant progress in recent years, showcasing impressive capabilities in processing and understanding both visual and textual information. Open-source models like the LLaVA series~\cite{liu2023llava, liu2024llavanext}, the Qwen-VL series~\cite{QwenVL, Qwen2VL}, and the InternVL series~\cite{internvl, internvl2_5} have demonstrated state-of-the-art performance across a variety of vision-language tasks. However, these models are primarily trained on large-scale third-person datasets, making them less effective in the egocentric video domain.
With the introduction of the Ego4D \cite{ego4d} dataset, several studies have made strides toward addressing this gap. For instance, EgoVLP \cite{egovlp} utilizes contrastive learning for egocentric video and language pretraining, while EgoVLPv2 \cite{pramanick2023egovlpv2} enables cross-model fusion in video and language backbones. LaViLa \cite{lavila} refines the pre-training data using large language models to learn egocentric video and language representations. In addition, the champion solution of InternVideo-Ego4D~\cite{chen2022internvideo-ego4d} ensembles video-only and video-language features to significantly enhance the performance across multiple egocentric downstream tasks.

While these models have made significant advancements, they are not optimized for real-time applications or continuous interaction. Vinci's EgoVideo-VL bridges this gap by connecting the egocentric foundation model EgoVideo to a large language model (LLM), enabling free-form language interaction. This integration allows Vinci to process long egocentric video streams in real-time while leveraging the LLM for flexible, natural language-based communication. This capability distinguishes Vinci from existing models, making it uniquely suited for dynamic, real-world egocentric AI applications.

\subsection{Smart Assistants on Portable Devices}
Smart assistants tailored for portable devices have gained significant traction. These systems aim to provide real-time, context-aware support by leveraging advancements in artificial intelligence, sensor integration, and mobile computing~\cite{guo2018device,tabassum2019investigating,gan2021year,adaimi2021ok,bentley2018understanding,zhou2022are,boldu2020aisee,chen2022lisee,li2022towards,an2021viscocam,paredes2022stretch,maruri2018vspeech,zhao2022do,janaka2024pilot}. Traditional virtual assistants such as Apple’s Siri~\cite{siri} and Google Assistant~\cite{googeassistant} provide speech-based interaction for tasks like scheduling, information retrieval, and basic navigation. These systems are primarily designed for smartphones or smart home devices and rely on cloud-based natural language processing models. More recently, several works~\cite{chen2024enabling,zulfikar2024memoro,arakawa2024prism,yang2024talk2care} leveraged LLMs for better language-based interactions. However, their limited integration with on-device visual inputs makes them less suitable for tasks requiring a deep understanding of real-world environments. 

With the recent development of Vision-language models, the integration of vision has marked a significant leap in the functionality of smart assistants. Many systems~\cite{yang2024socialmind,jing2024anglesizer,xu2024can,wen2024find,pataranutaporn2023living,yang2024drhouse} extend traditional assistants by incorporating image recognition and video analysis capabilities, allowing for more advanced functionalities such as object identification, scene description, and visual question answering. However, these assistants are often designed for sensing the outside environment, lacking the ability to adapt to the dynamic, egocentric nature of video captured from wearable devices. Since the egocentric camera wearer is often invisible or only partly visible, as we will show in our experiments, most general models cannot satisfactorily perceive the egocentric actions of the camera wearer.

\subsection{Egocentric Computer Vision}
The egocentric perspective introduces unique challenges in human activity analysis~\cite{plizzari2023outlook}, requiring specialized techniques for understanding interactions, predicting future actions, and reasoning about pose in first-person views~\cite{bock2024wear,mahmud2023posesonic}. Core focus in egocentric computer vision include action recognition and detection~\cite{girdhar2021anticipative,wang2023ego,plizzari2022e2,radevski2023multimodal}, hand detection~\cite{shan2020understanding,zhang2022fine,goyal2022human}, video-language understanding~\cite{kang2021video,huang2023weakly,xu2024retrieval}, and gaze estimation~\cite{huang2018predicting,huang2020mutual,zhao2023unveiling,chong2017detecting}. Egocentric vision research has broad implications for embodied AI~\cite{nagarajan2021shaping}, where systems must process long-term observations and reason over sequential experiences to plan actions effectively. It also plays a key role in VR/AR~\cite{liu2022joint,Narayanan2024retro,yan2020head,wang2024mr,hagan2023privacy,tian2019enhancing,wee2018focus}, where understanding user actions enhances immersive experiences and real-time interactions. Furthermore, egocentric video understanding contributes to human-machine interaction~\cite{marina2022head,xin2023learning,li2019deep,qi2023cases}, enabling more intuitive and responsive AI assistants.

In this work, we build upon these foundations to introduce Vinci, a real-time smart assistant for portable egocentric AI. Vinci integrates a vision-language model with a frontend-backend architecture optimized for continuous, real-world deployment. By combining state-of-the-art multimodal reasoning, memory-augmented processing, and low-latency inference, Vinci provides contextual and actionable insights in real-time. We envision Vinci as a step toward seamless, always-available egocentric AI, inspiring further advancements in smart assistive systems that naturally integrate into daily life.

\section{SURVEY ON SMART ASSISTANT NEEDS}
To better understand the user needs and expectations for egocentric smart portable assistants during daily life, we conduct a survey exploring user experience,
preferences, and needs. We use the results and findings as a guide for our system design. The goal of this survey is to gather insights into users' familiarity with smart wearable technologies, their desired functionalities, and their expectations for egocentric wearable assistants. This information is crucial for guiding the design and development of Vinci to align with user preferences and address potential usability challenges.

\subsection{Questionnaire}
The questionnaire was carefully designed to capture diverse aspects of user interaction and expectations with portable and wearable smart assistants. It comprised 10 questions, categorized into three sections: user familiarity and preferences, demographic information, and open-ended feedback:

\noindent\textbf{1) User Familiarity and Preferences:} This section, containing five multiple-choice questions, aims to assess users' prior experience with wearable assistants and their level of awareness of egocentric smart assistant technologies. It also explores users' desired features and the functionalities they find most appealing or essential.

\noindent\textbf{2) Demographic Information:} 4 questions on basic demographic data, including age, gender, race, and occupation, are used to understand how user needs and preferences might vary across different population groups.

\noindent\textbf{3) Open-ended Feedback:} The final section contains one open-ended question as an opportunity for participants to share their expectations for egocentric wearable assistants and offer additional recommendations or ideas for improvement.

In total, we collected 107 questionnaires from participants with diverse demographics and summarized the results and findings in the following section.

\begin{figure}
    \centering
    \includegraphics[width=\linewidth]{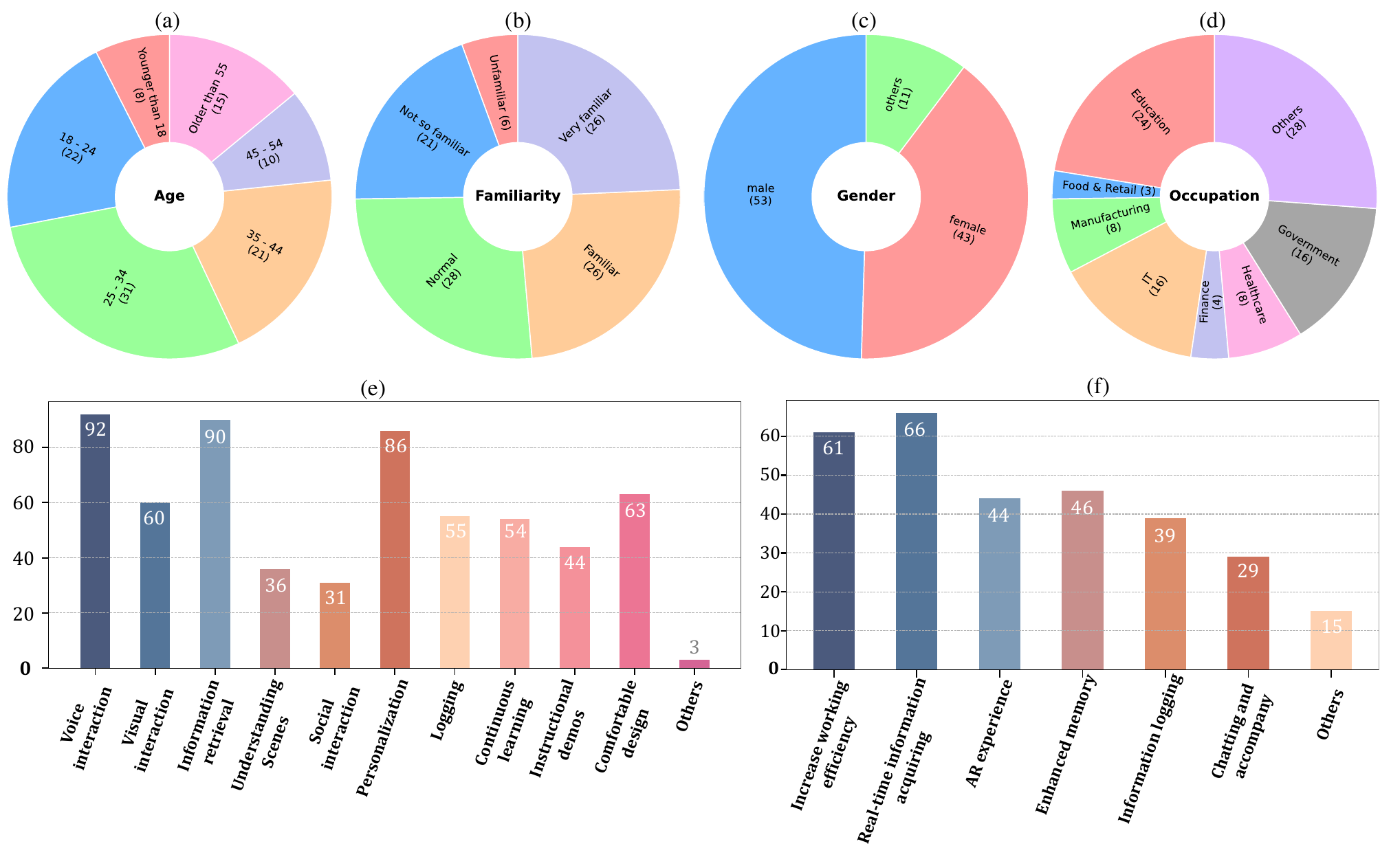}
    \caption{Results of the pre-experiment survey. Pie charts (a-d) show the demographics of the survey participants, where (a) is the age distribution, (b) is the familiarity with smart assistants, (c) is the gender, and (d) is the occupation. Bar plot (e) shows the participants' expectations of what features an egocentric smart assistant should have. Bar plot (f) is the result of the participants' thoughts on what aspect of an egocentric smart assistant is the most appealing.}
    \label{fig:pre}
\end{figure}

\subsection{The Demand and Preference for an Egocentric Smart Assistant}
From Figure~\ref{fig:pre}, several insights can be drawn regarding user preferences and expectations for egocentric wearable smart assistants:

The three most important basic functions identified were voice interaction, information retrieval, and personalization. Over half of the participants emphasized the importance of a comfortable design. Secondary needs included visual interaction, logging, continual learning, and instructional demos. Features related to scene understanding and social interactions were considered less critical by the participants.

When asked about the most appealing aspects of a first-person wearable smart assistant, participants prioritized features that increase working efficiency and enable real-time information acquisition. These results highlight a strong preference for practical and time-saving functionalities.

Based on the survey results, we believe the following features are essential for an egocentric portable smart assistant:

1. \textbf{Voice Interaction:} The assistant should enable intuitive and responsive voice-based commands to facilitate hands-free usage. 

2. \textbf{Real-time response:} The assistant must minimize latency to ensure timely and seamless interaction.

3. \textbf{Comfortable Design:} Ergonomic and lightweight form factors are crucial for ensuring user comfort during extended use.

4. \textbf{Personalization and continual learning:} The ability to adapt gradually to personalized user habits and preferences, enhancing functionality over time.

5. \textbf{Information Retrieval:} Real-time access to relevant and contextual information tailored to the user’s action and instruction.

6. \textbf{Instructional Demos:} Detailed and visualized guidance for ongoing tasks, dynamically adapted to the user’s environment.

These findings serve as guiding principles in the development of Vinci, ensuring it meets user expectations and needs effectively.

\section{SYSTEM DESIGN}
\subsection{System Overview}
\begin{figure}
    \centering
    \includegraphics[width=\linewidth]{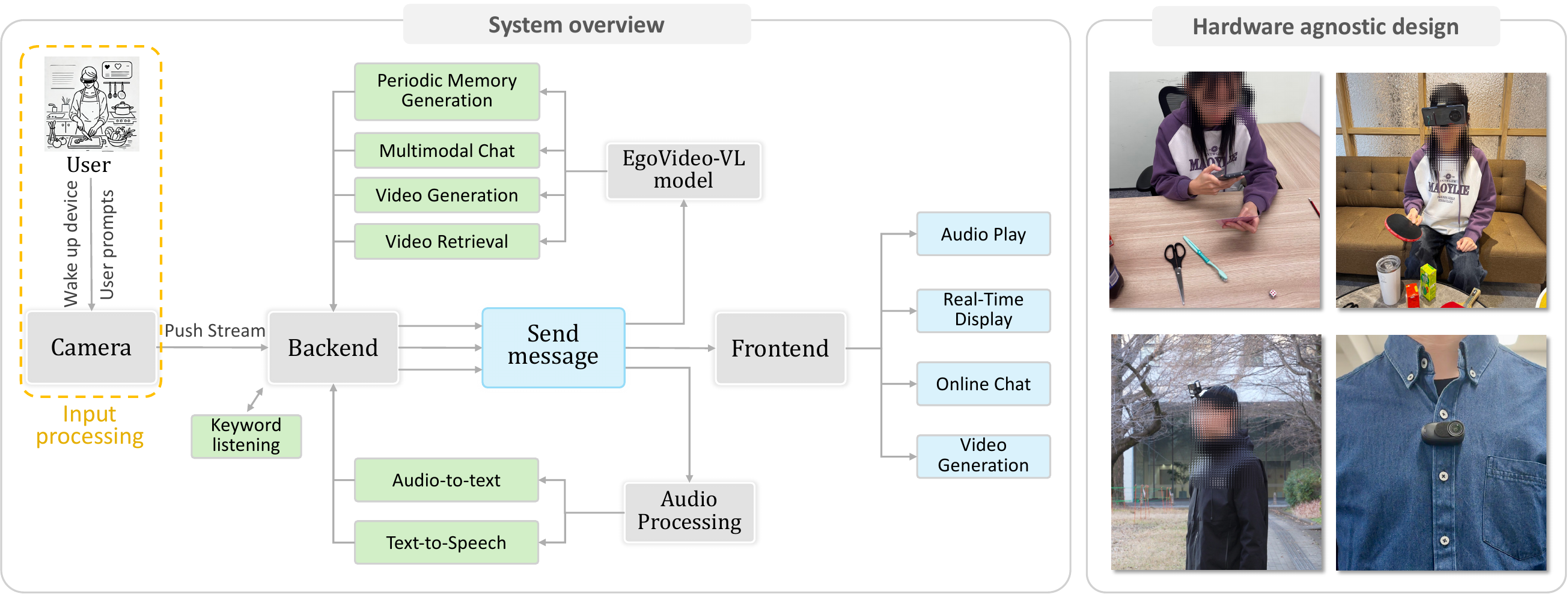}
    \caption{\textbf{Overview of the Vinci system.} The left side illustrates Vinci’s system architecture, comprising four key components: (1) Input Processing Module, which receives live video streams and user queries; (2) Backend, which manages communication, query processing, and wake-up keyword detection; (3) EgoVideo-VL Model, which integrates egocentric vision with language understanding for real-time multimodal reasoning; and (4) Frontend, which delivers responses via text, speech, or visual demonstrations. The right side shows examples of real-world hardware deployments, demonstrating Vinci’s versatility across smartphones and wearable cameras for seamless, context-aware assistance in dynamic environments.}
    \label{fig:system}
\end{figure}

Vinci is a real-time egocentric vision-language assistant designed for deployment on portable devices such as smart glasses, GoPro cameras, and smartphones. It also supports online inference and interaction on pre-recorded videos in any format. \textit{A hardware-agnostic design is chosen to allow users to utilize the devices they find most comfortable, aligning with the needs identified in the previous section.} An overview of the system's workflow is depicted in Figure~\ref{fig:system}.
The system is composed of four primary modules:

\noindent\textbf{1) An Input Processing Module}, which receives live-streaming video and audio, while also transcribing audio into text. This module is essential in addressing the users' need for audio-based interaction. 

\noindent\textbf{2) A Vision-Language Model EgoVideo-VL} that processes visual and textual inputs, interprets user queries, and generates responses. 
We choose a vision-language model over a language-only model because it leverages both visual and textual information, providing a more comprehensive understanding of the user's environment. In egocentric applications, visual cues such as objects, actions, and spatial layouts are crucial for accurate interpretation and effective decision-making.

To better align with users' needs, we have meticulously designed and trained our EgoVideo-VL model. This process involves aligning the state-of-the-art egocentric foundation model with a large language model (LLM) and curating additional data to further enhance the alignment. 

We also incorporate specialized modules tailored to address specific user requirements.
Within EgoVideo-VL there is a Memory Module, which integrates and retrieves historical context to maintain continuity and personalized interaction experience. Also, a generation module produces instructional video demonstrations to guide the user. Meanwhile, a retrieval module aims to retrieve relevant videos from a database based on the user's requirements. 

\noindent\textbf{3) A backend} hosting HTTP services of the input processing module and the model services; 

\noindent\textbf{4) A web-based frontend}, which shows the model output and plays the audio generated by the text-to-speech (TTS) technique. These modules operate in parallel, enabling Vinci to deliver seamless, real-time assistance, ensuring robust performance and effective user support. Below, we go into the details of each module.

\subsection{Input processing}
The input processing module prepares live-streaming data for downstream processing and supports any RTMP-compatible device, enabling versatility across platforms. Vinci accommodates various camera setups, including: 1) Smartphones equipped with streaming apps, 2) Wearable cameras such as GoPro, and 3) Standard webcams configured with FFmpeg for streaming. In our experiments, Vinci has been successfully deployed on iPhones, Android smartphones, GoPro cameras, DJI cameras, and webcams, demonstrating its adaptability across diverse hardware.

The input processing module handles live-streaming video and audio inputs, with audio transcribed into text using an automatic speech recognition (ASR) system. For our open-source demonstration and in our experiments, we use Baidu’s ASR API, but the system is modular and can integrate alternative ASR solutions as needed. In parallel, the video stream is processed to synchronize with the transcribed audio, maintaining alignment between visual and textual inputs. This capability ensures that all incoming data is structured and ready for real-time analysis by the vision-language model. 

\begin{figure}
    \centering
    \includegraphics[width=\linewidth]{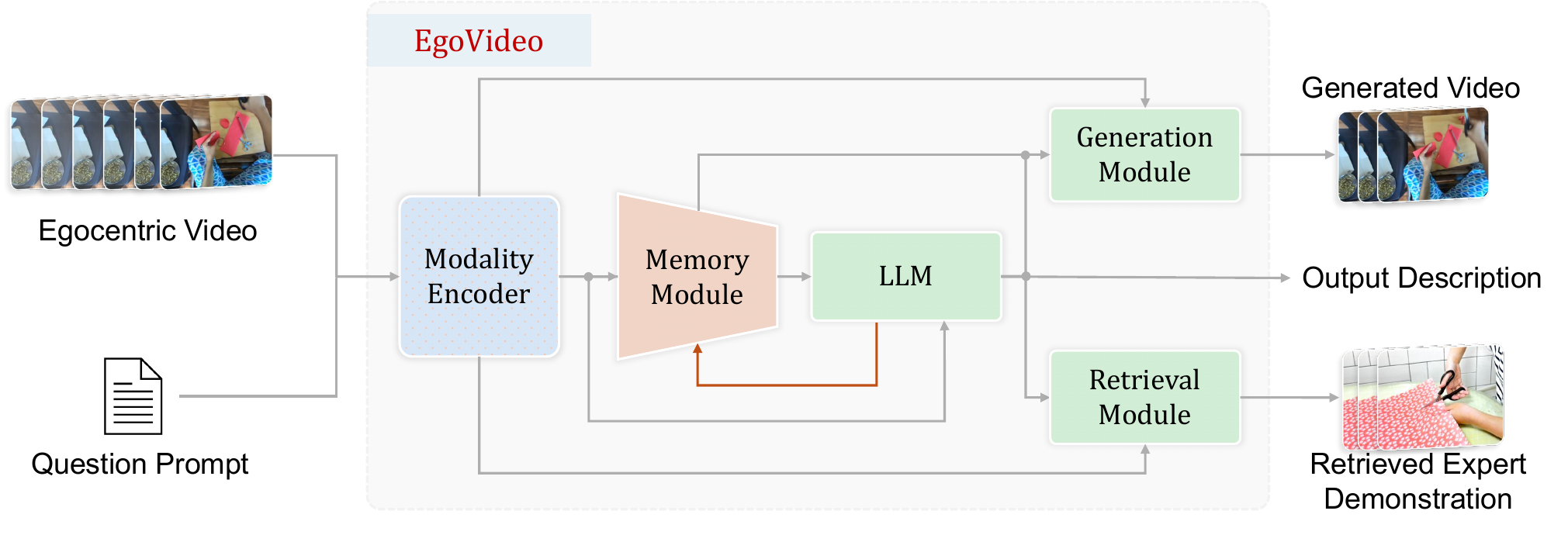}
    \caption{\textbf{Overview of the EgoVideo-VL model.} EgoVideo-VL is a multimodal vision-language model designed for real-time egocentric understanding and assistance. The model comprises five key components: (1) Modality Encoder, which follows the design of EgoVideo~\cite{pei2024egovideo} and includes a video encoder and a text encoder for multimodal feature extraction; (2) Memory Module, which stores historical context to enable temporal grounding, summarization, and personalized interactions; (3) Large Language Model (LLM), which performs multimodal reasoning and response generation; (4) Generation Module, which synthesizes visual action predictions to guide users through tasks; and (5) Retrieval Module, which retrieves third-person expert demonstrations to complement egocentric understanding.}
    \label{fig:vl}
\end{figure}

\subsection{EgoVideo-VL}
At the core of Vinci’s functionality lies its vision-language model (VLM), which processes the video and text input from the input processing module. An overview of EgoVideo-VL is presented in Figure~\ref{fig:vl}.
Unlike existing vision-language models such as LLaVa or InternVL, which do not account for the “ego” perspective, we designed EgoVideo-VL to address this gap. A key requirement for portable assistants is the ability to focus on the needs of the user, who is often not directly visible to the camera. To meet this requirement, we propose EgoVideo-VL, a vision-language model that leverages the strong egocentric video understanding capabilities of the state-of-the-art foundation model EgoVideo~\cite{pei2024egovideo}. EgoVideo excels in various egocentric video tasks, such as action recognition and anticipation, where the focus is on the actions of the camera wearer. Below, we describe how we integrate these capabilities into EgoVideo-VL and extend its functionality to support real-world human needs.

\subsubsection{Vision-language model design.} 
EgoVideo $\phi_v$ is a Transformer-based video encoder~\cite{vit}. 
It takes as input a video $V\in\mathbb{R}^{T\times H\times W\times 3}$ of $T$ frames and $H\times W$ spatial resolution,  converts it into video patches, and encodes it with stacked Transformer Encoder layers.
The output of EgoVideo $f_v = \phi_v(V)\in\mathbb{R}^{n\times d}$ where $n$ refers to the number of spatiotemporal patches and $d$ is the hidden dimension. 
We connect the visual token outputs of EgoVideo to a large language model (LLM) to seamlessly integrate visual and textual information. 
Specifically, the text instruction is tokenized and transformed into text embeddings $I\in\mathbb{R}^{L\times d}$ with sequence length $L$. 
The input to the LLM consists of special <image> tokens representing visual content followed by visual and text embeddings, enabling joint processing of multimodal data. By default, the vision-language model outputs the response $R$ to the user's query. 

\subsubsection{Data Curation.} 
To enhance EgoVideo-VL's ability to handle long-form egocentric videos and perform tasks such as temporal grounding and future planning, we curate a specialized dataset for training. 
This dataset combines data from three sources: Ego4D~\cite{ego4d}, EgoExoLearn~\cite{huang2024egoexolearn}, and Ego4D-Goalstep~\cite{song2024ego4d}. Specifically, we use the narrations from Ego4D and EgoExoLearn to create 4 million video-text pairs. For each video-text pair, we assign task-specific prompts drawn from a set of LLM-refined templates, forming video-instruction-answer triplets for training.

For Ego4D-Goalstep, which provides procedural step and substep annotations, we generate data that focuses on procedural task planning and historical reasoning. In this context, we treat previous actions as memory and incorporate this into the model’s training. Specifically, we modify the input format from "<instruction>" to "<structured memory>, <instruction>", creating video-memory-instruction-answer quadruplets. This enables the model to leverage historical context for improved task planning. Further details on the structured memory are described in the next section.

\subsubsection{Memory integration.} 
Due to the computational challenges of processing all streaming video frames, we introduce a memory module $\mathcal{M}$ to retain historical context and improve the personalized experience of the Vinci system. The memory module plays a crucial role in handling the continuous and streaming nature of egocentric data. 
In particular, the memory module operates in a store-and-update manner. 
At the storing stage, it leverages the vision-language model to capture short video snapshots $V_i$ at timestamp $s_i$ by generating textual descriptions of observed actions $h_i$, \emph{e.g.} pour the water into the pot.
The textual description, along with the corresponding timestamps $i$ is stored into the structured memory bank $\mathcal{M}=[(h_1,s_1),(h_2,s_2)...,(h_{i-1},s_{i-1}),(h_i,s_i)]$, where $[\cdot]$ denotes concatenation.
The memory module stores the historical context in a FIFO (First-In-First-Out) queue, where the context length $N_{\mathcal{M}}$ can be specified by the user based on his/her own computation resources. 
Whenever the memory module exceeds the maximum context length, it is updated by discarding the 
earliest context, \emph{i.e.},
\begin{equation}
 \mathcal{M}=[(h_{i-N_{\mathcal{M}}+1},s_{i-N_{\mathcal{M}}+1}),...,(h_{i-1},s_{i-1}),(h_i,s_i)]   
\end{equation}
In this way, this memory bank ensures that the system maintains a comprehensive record of past events. 
When a user interacts with Vinci, the memory module provides relevant historical context to EgoVideo-VL, allowing the system to temporally ground past actions, summarize user activities, and answer queries requiring historical reasoning. 
By integrating this memory framework, Vinci achieves robust understanding of temporal sequences, enabling advanced functionalities such as multi-step task planning and natural language grounding.

\subsubsection{Instruction fine-tuning.} To train EgoVideo-VL, we employ the LoRA (Low-Rank Adaptation) technique for fine-tuning. The fine-tuning is conducted in two stages:
\begin{enumerate}
    \item In the first stage, we fine-tune the model using data only from Ego4D and EgoExoLearn, to align the vision and language tokens in the context of egocentric video.
    \item In the second stage, we incorporate data from all three datasets, with a focus on Ego4D-Goalstep data, which is designed to improve task planning and historical reasoning. During training, the LLM component (InternLM-7B~\cite{cai2024internlm2}) remains fixed, providing a stable foundation for language understanding and generation.
\end{enumerate}

\subsubsection{Generation model integration.}
The generation module in Vinci is responsible for creating visual how-to demonstrations, enabling users to understand and perform tasks that require detailed visual guidance. 
This module is built upon SEINE~\cite{seine}, an image-to-video generation model from the open-source video generation system Vchitect~\cite{lavie,seine,latte,vbench,vbench++++}, which we fine-tune to address the specific challenges of egocentric video generation.

To fine-tune SEINE, we curate a specialized subset of the instruction-tuning dataset, selecting videos based on two key criteria: 
(1) videos must exhibit smooth global motion, which we ensure by selecting those with a maximum optical flow below a defined threshold, 
and (2) the verbs in the associated text must occur with reasonable frequency, preventing the model from collapsing during training due to rare verbs.

The generation model $\phi_{gen}=\{\mathcal{E},\epsilon_{\theta},\mathcal{D}\}$ consists of a VAE encoder $\mathcal{E}$, a denoising UNet $\epsilon_{\theta}$ parameterized by $\theta$, and a VAE decoder $\mathcal{D}$.
At training time, given a video $V$ with $T$ frames, it is first encoded by the frozen VAE encoder $z_0=\mathcal{E}(V)\in\mathbb{R}^{T\times h\times w\times c}$ in a frame-by-frame manner. Following the standard diffusion process~\cite{seine,latte,lavie}, the encoded video feature $z_0$ is corrupted by adding noise $\epsilon \sim \mathcal{N}(0, \mathbf{I})$, where $\mathcal{N}(0, \mathbf{I})$ refers to the standard Gaussian distribution. We denote the corrupted video feature as $z_t$ where $t$ here denotes the diffusion time-step. 
To make sure the current observed context serves as the structural guidance for producing reliable how-to demonstration, the input to the model $\overline{z}$ is the channel-wise concatenation of three components. (1) the corrupted video feature $z_t$; (2) the first frame of the un-corrupted video feature $z_0^1$; (3)
a binary temporal mask [1, 0, ..., 0] of length $T$, indicating that the first frame is given as the context.
We fine-tune the denoising U-Net $\epsilon_{\theta}$ using the standard $\epsilon$-prediction loss. 
\begin{equation}
    \mathcal{L}_{\text{gen}} = \mathbb{E}_{t,V\sim p_{\text{data}},\epsilon\sim \mathcal{N}(0,\textbf{I})} \|\epsilon - \epsilon_{\theta}(\overline{z},I,t)\|_2^2, 
\end{equation}
where $I$ is the user's instruction.

During inference, EgoVideo-VL first determines whether generation is necessary based on the user’s query. If required, the system provides the query $I$ and the most recent video frame to the generation module, which uses the DDIM~\cite{song2020denoising} sampling with 50 steps and outputs a 2-second video demonstrating the requested action. This functionality empowers Vinci to provide actionable, visual guidance tailored to the user’s needs.

\subsubsection{Retrieval model integration.}
The retrieval module in Vinci aims to automatically retrieve how-to videos based on the user’s query, providing visual demonstrations and references for completing ongoing tasks. Unlike the generation module, which directly generates visual guidance using a video diffusion model, the retrieval module sources videos from a large database of instructional content, such as HowTo100M~\cite{howto100m}.

To implement this, we leverage the retrieval module in EgoInstructor~\cite{xu2024retrieval}, a model trained on pseudo-paired egocentric-exocentric videos~\cite{ego4d,howto100m}. 
The retrieval module consists of a text encoder $\phi_{r-text}$ and a paired visual encoder $\phi_{r-vis}$.
Additionally, Vinci maintains a database containing a vast amount of how-to demonstration videos for the retrieval module. 
Before deployment, we encode each how-to demonstration video $V$ with the trained visual encoder, \emph{i.e.,} $f_v=\phi_{r-vis}(V)$.
We store all the extracted video features on the disk, \emph{i.e.,} $\mathcal{F}=\{f_{v_1},f_{v_2},...,f_{v_N}\}$.

Given a user query $I$ (e.g., "show me how to cut a tomato"), the text encoder first extracts text features of the query $f_t=\phi_{r-text}(I)$. Vinci then computes pairwise cosine similarity $<f_v,f_t>$ with all the video features in $\mathcal{F}$. 
The top-K most similar videos are then retrieved using FAISS~\footnote{\url{https://github.com/facebookresearch/faiss}}. 
\begin{equation}
    V_i = \text{argmax}_{i}<f_{v_i},f_t>
\end{equation}
In our system, we set $K$=3 to retrieve the three most relevant videos. This allows Vinci to provide visual guidance sourced from a vast range of online instructional videos, reducing the need for users to manually search platforms like Google or YouTube.

\subsection{Frontend and backend}
The Vinci system is composed of four primary components: the Input Processing Module, the Vision-Language Model (EgoVideo-VL), the Frontend, and the Backend. These modules work together in parallel to provide seamless, real-time egocentric AI assistance. 

\begin{figure}
    \centering
    \includegraphics[width=0.9\linewidth]{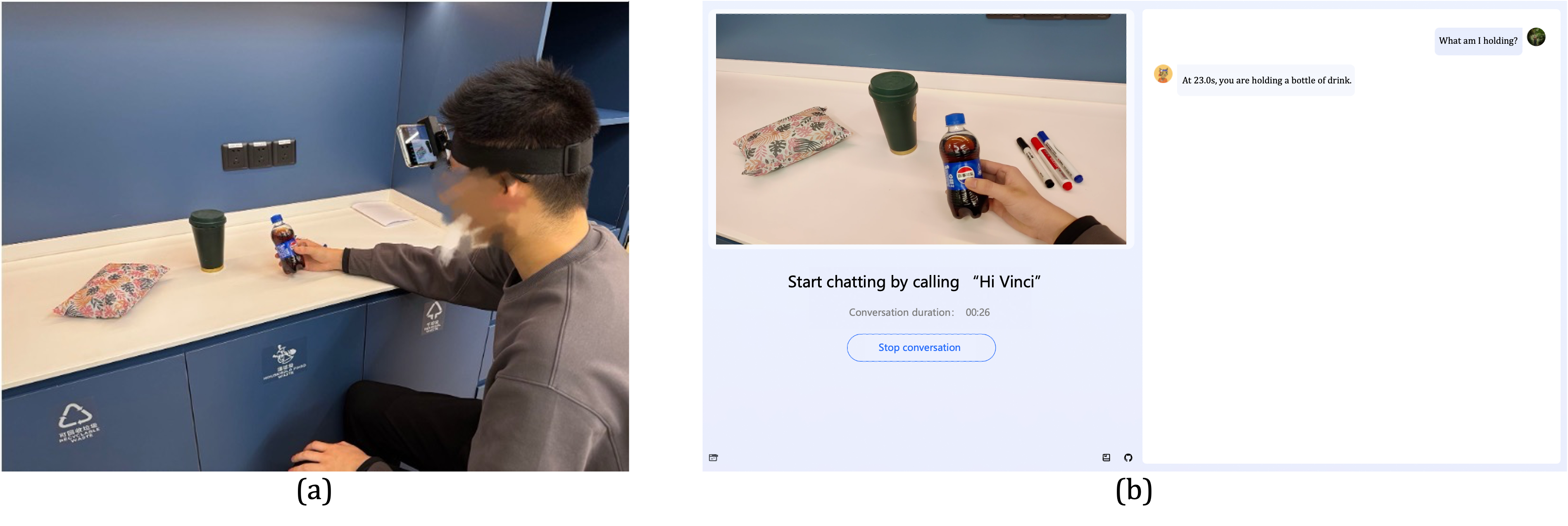}
    \caption{(a) Example of Vinci operating on a head-mounted smartphone.  (b) The web-based user interface of Vinci, displaying the live video stream, conversation history, and audio playback.}
    \label{fig:realworld}
\end{figure}

The Frontend is a web-based interface powered by JavaScript. It establishes a WebSocket connection with the Backend to receive and display video and text outputs. The Frontend continuously shows the live-streaming video. When a text response from EgoVideo-VL is received, the system uses a Text-to-Speech (TTS) engine to convert the text into speech and plays the audio. If video data is generated by the Generation Module or retrieved by the Retrieval Module, the Frontend plays the video accordingly. This ensures smooth interaction, with the Frontend visually displaying the system’s responses and providing a real-time interface for user interaction. An example of operating Vinci and its user interface can be found in Figure~\ref{fig:realworld}.

The backend serves as the core computational engine of the Vinci system, orchestrating real-time processing and communication between components. It hosts HTTP services that facilitate seamless interaction between the Input Processing Module, the EgoVideo-VL model, and the Frontend, ensuring efficient data flow and low-latency responses. By decoupling input processing from model execution, the backend enables parallel processing, maximizing GPU utilization while maintaining real-time performance. 

A key functionality of the backend is wake-word detection, implemented via an API-based solution~\cite{wenwen}. When the user-defined wake-up keyword is detected in the audio stream, the backend triggers the processing pipeline by forwarding the user's query and the most recent 2-second video snippet to the Input Processing Module. To handle multiple concurrent interactions, an adaptive queuing mechanism ensures sequential query processing, preventing conflicts when new queries arrive before prior ones are resolved.

Upon generating a response, the EgoVideo-VL model transmits its output to the backend, which then synchronizes it with the live video stream before relaying both to the Frontend. This design ensures fluid user interaction, allowing Vinci to deliver accurate, context-aware responses with minimal latency, which is critical for real-world deployment on mobile and wearable platforms.

\section{EVALUATION}
This section introduces the experimental setup, evaluation of Vinci, and a real-world user study.

\subsection{Performance of the core model: EgoVideo-VL}
First, we evaluate our EgoVideo-VL on egocentric video understanding benchmarks, to validate its superiority against other vision-language models on smart assistants that take the egocentric view. We use the following benchmark tasks:
\begin{itemize}
    \item \textbf{EK-100 MIR:}This task aims to develop models capable of retrieving relevant video segments from the Epic-Kitchens-100 dataset based on a textual query describing an action or activity. The evaluation metrics include Mean Average Precision (mAP) and normalized Discounted Cumulative Gain (nDCG). For more details, please refer to~\cite{damen2018scaling}.
    \item \textbf{EGTEA:} We evaluate the zero-shot action recognition accuracy on the EGTEA dataset~\cite{li2018eye}, which comprises fine-grained cooking activities. The evaluation metrics are Mean Class Accuracy and Top-1 Accuracy, capturing the model's ability to recognize actions without task-specific fine-tuning.
    \item \textbf{EgoMCQ:} This is a benchmark of multiple choice questions \cite{lin2022egocentric} focuses on evaluating egocentric video-text alignment. The data are selected from Ego4D, and the evaluation metrics are MCQ accuracies on 1) "inter-video" questions, where the choice options originate from different videos, and 2) "intra-video" questions, where the choice options come from the same video.
    \item \textbf{EgoSchema:} This benchmark evaluates long-term video understanding in egocentric videos. Each multiple-choice question requires reasoning over extended temporal contexts to arrive at the correct answer. Unlike shorter video datasets, EgoSchema challenges models with complex, temporally extended scenarios. 
\end{itemize}

\renewcommand{\arraystretch}{1.2}
\begin{table}[t!]

\caption{Performance comparison of Vinci's core model EgoVideo-VL and other models on different benchmarks.} 
\label{tab:model}
\centering
\resizebox{0.9\linewidth}{!}{
% \vspace{-0.1in}
    \begin{tabular}{lc|cc|cc|cc|c}
 \Xhline{1.0pt}
    \rowcolor{mygray}
     & & \multicolumn{2}{c|}{\bf EK-100 MIR} &  \multicolumn{2}{c|}{\bf EGTEA} & \multicolumn{2}{c|}{\bf EgoMCQ} &  \\
     \rowcolor{mygray} \multirow{-2}{*}{\bf Model} &\multirow{-2}{*}{\bf LLM}  & \textit{mAP} & \textit{nDCG} & \textit{Mean-acc.} & \textit{Top1-acc.} & \textit{Intra} & \textit{Inter} & \multirow{-2}{*}{\bf EgoSchema} \\
\Xhline{0.7pt}
InternVideo~\cite{chen2022internvideo-ego4d} & - & 34.7 &49.7 &33.5& 39.3& 50.7 &90.4 & 32.1 \\
LaViLa~\cite{zhao2023learning} & - & 36.1 & 34.6 &34.1 &40.1 &60.9& 91.0& 23.5  \\
EMBED~\cite{dou2024unlocking} & - & 40.8 &37.5 &40.3& 46.7& 64.7 &95.6 &- \\
\Xhline{0.7pt}
TimeChat~\cite{ren2024timechat} &  LLaMA2-7B~\cite{touvron2023llama} & - &- &-& -& - &- &33.0 \\
ReCap~\cite{islam2024video} & GPT3.5~\cite{gpt4o} & - &- &-& -& - &- &50.2 \\
InternVideo2~\cite{wang2024internvideo2} & Mistrial-7B~\cite{jiang2023mistral} & 39.5 &36.0 & 43.1 & 49.7& 64.5& 95.2 &55.8 \\
\bf EgoVideo-VL & InternLM-7B~\cite{cai2024internlm2} & \bf 47.1 & \bf 39.0 & \bf 58.0 & \bf 63.0 & \bf 69.1 & \bf 96.6 & \bf 60.2  \\ 
 \Xhline{1.0pt}
    \end{tabular}
}
\end{table}

For comparison, we use two groups of models. The first group of models are egocentric video understanding models without LLM. Within this group, in addition to a general video understanding model InternVideo~\cite{chen2022internvideo-ego4d}, we specifically compare with the state-of-the-art models designed specifically for egocentric video understanding tasks.  We use the state-of-the-art models LaViLa~\cite{zhao2023learning} and EMBED~\cite{dou2024unlocking} for comparison. While these models excel in processing egocentric video data, they lack the capability for human-computer interaction due to the absence of a language model. Another group of models we compare are general vision-language models. This group consists of models that incorporate large language models (LLMs), enabling them to perform language-based interactions. However, they are not specifically tailored for egocentric tasks and may struggle with understanding first-person perspectives. We select three representative models for this category: TimeChat~\cite{ren2024timechat}, ReCap~\cite{islam2024video}, and InternVideo2~\cite{wang2024internvideo2}, all of which are advanced video-based vision-language models that serve as strong baselines for comparison with our approach.

The results can be seen in Table~\ref{tab:model}.
The general video understanding model InternVideo cannot perform well on all benchmarks, since it is not specially trained on egocentric data.
For the egocentric video understanding models, such as LaViLa and EMBED, both demonstrate strong performance on the traditional benchmarks (EK-100 MIR, EGTEA, and EgoMCQ), which focus on core egocentric video understanding tasks. However, these models perform poorly on the EgoSchema benchmark due to the absence of LLMs. EgoSchema requires not only free-form language understanding but also the ability to aggregate information over extended video sequences. Without LLM, these capabilities are lacking in the video understanding models.

In contrast, vision-language models like ReCap and InternVideo2 show significantly better performance on EgoSchema due to their LLM-based language processing capabilities. However, their performance on traditional egocentric video understanding benchmarks is comparatively weaker, for instance, InternVideo2 even falls short of EMBED. This suggests that while these models are better at describing "what happens in the scene," they struggle to understand "what the camera wearer is doing," a critical aspect for egocentric applications.

On the other hand, our EgoVideo-VL performs well on all these benchmarks. 
It achieves competitive results on EK-100 MIR, EGTEA, and EgoMCQ, highlighting its strength in egocentric video understanding. Simultaneously, it excels on EgoSchema, showcasing its ability to perform natural language-based interactions and process long-term video information. 
This dual strength validates our design and training approach for EgoVideo-VL, making it a robust choice as the core model in Vinci for egocentric smart assistant applications. 

With these results, we proceed to the next section, where we conduct multiple studies to evaluate the value of each functionality of Vinci and assess its performance from a human-centric perspective.

\subsection{Evaluation on each functionality}
In this section, we present a series of experiments including controlled experiments and in-situ user studies, designed to evaluate each functionality offered by Vinci. For each functionality, we outline the experimental setup and employ specific metrics for validation. The details of each experiment are provided below.

\subsubsection{Contextual Chatting}
The chatting functionality represents Vinci's overall performance. In this evaluation, we assess Vinci's ability to handle real-time, context-aware, egocentric conversations and evaluate its response speed, through controlled experiments and an in-situ user study.

\begin{table}[t!]

\caption{Controlled test on Vinci's performance on Contextual Chatting. We evaluate whether Vinci can correctly recognize the ongoing action from both verb and noun perspectives.} 
\label{tab:chat}
\centering
\resizebox{\linewidth}{!}{
% \vspace{-0.1in}
    \begin{tabular}{l|cccccccccc}
    \Xhline{1.0pt}
    \rowcolor{mygray} Accuracy & Pen & Pencil & Scissors & Cup & Umbrella & Toy & Mouse & Calculator & Toothbrush & Cards \\
\Xhline{0.7pt}
% Noun & 100\% & 90\% & 95\% & 95\% & 80\% & 95\% & 100\% & 85\% & 95\% & 95\% \\
Put & 90\% & 90\% & 80\% & 85\% & 90\% & 90\% & 95\% & 85\% & 90\% & 85\% \\
Take & 85\% & 85\% & 80\% & 90\% & 85\% & 90\% & 90\% & 80\% & 85\% & 85\% \\
Operate & 90\% & 90\% & 85\% & 85\% & 85\% & 80\% & 85\% & 95\% & 90\% & 90\% \\
Hold & 100\% & 100\% & 100\% & 100\% & 95\% & 95\% & 80\% & 90\% & 95\% & 95\% \\
Rotate & 95\% & 100\% & 100\% & 100\% & 90\% & 100\% & 90\% & 85\% & 100\% & 85\% \\
 \Xhline{1.0pt}
    \end{tabular}
}
\end{table}

\noindent\textbf{Controlled evaluation.$\quad$} 
For controlled evaluation, we selected 10 types of daily-use objects and performed human-object interactions with them. The interaction types included take, put, hold, operate, and rotate. We evaluated Vinci's performance in terms of object recognition accuracy and action recognition accuracy. 
We do not use existing action recognition datasets like EK-100 or EGTEA, since the model's performance on these datasets was already validated in the previous section. Thus, we used newly collected, unseen data for this evaluation, better demonstrating the real-world application ability of Vinci in more challenging new, zero-shot scenarios. This can better assess Vinci's zero-shot generalization in real-world scenarios.

\noindent\textbf{Results.$\quad$} As shown in Table~\ref{tab:chat}, Vinci achieved approximately 90\% recognition accuracy across all ten object categories: pen, pencil, scissors, cup, umbrella, toy, mouse, calculator, toothbrush, and cards. The lowest accuracy was observed for umbrella (sometimes misclassified as a bag) and calculator (misclassified as a remote), primarily due to visual similarities in a zero-shot setting. For action recognition, Vinci achieved at least 80\% accuracy across all interaction types (verbs), with particularly strong performance on dynamic actions like "rotate" due to its video-based processing. These results highlight Vinci’s capability in recognizing both objects and actions in diverse environments.

\noindent\textbf{In-Situ user study.$\quad$}
Beyond controlled tests, we also conduct an in-situ user study to evaluate Vinci's Chatting ability comprehensively. For this user study, we recruited 20 participants from diverse backgrounds to interact with Vinci in two real-world environments: indoor daily activities and outdoor train station navigation.
In the indoor daily activities setting, participants performed everyday tasks, such as cooking, fitness training, and desktop working.
In the train station setting, participants navigated through the station, seeking directions to platforms, ticket counters, or exits while interacting with Vinci for real-time assistance.
In both scenarios, each participant recorded videos ranging from 5 minutes to 20 minutes, and during which time each participant will interact with Vinci 5 times.
For hardware configurations in the free-form user study, 15 participants performed indoor tasks and used smartphones held by hand or worn on the head using a band, and 5 participants performed outdoor experiments and used head or chest-mounted wearable cameras. We will omit the description of this setup in the remaining sections of this paper since all other in-situ experiments follow the same setting. 

For evaluating the chatting functionality, we do not use predefined tasks, but ask the participants to naturally interact with Vinci. After that, we save all queries and responses, and ask the participant to evaluate from two perspectives: accuracy and satisfaction score. We also compute the response speed based on the system log. Accuracy is defined as the percentage of correct responses compared to the participant's expectations. The user satisfaction score: is collected via post-interaction surveys, rated on a 5-point Likert scale, assessing clarity, relevance, and overall user experience.

\noindent\textbf{Results.$\quad$} Quantitative results can be found in the first row of Table~\ref{tab:user}. Vinci achieved the following performance:
The accuracy reached 91.0\% in indoor tests and 84.0\% in outdoor tests, demonstrating strong contextual and conversational understanding. The average response time was 0.7 seconds, with a standard deviation of 0.3 seconds, ensuring real-time usability. 

\begin{figure}
    \centering
    \includegraphics[width=\linewidth]{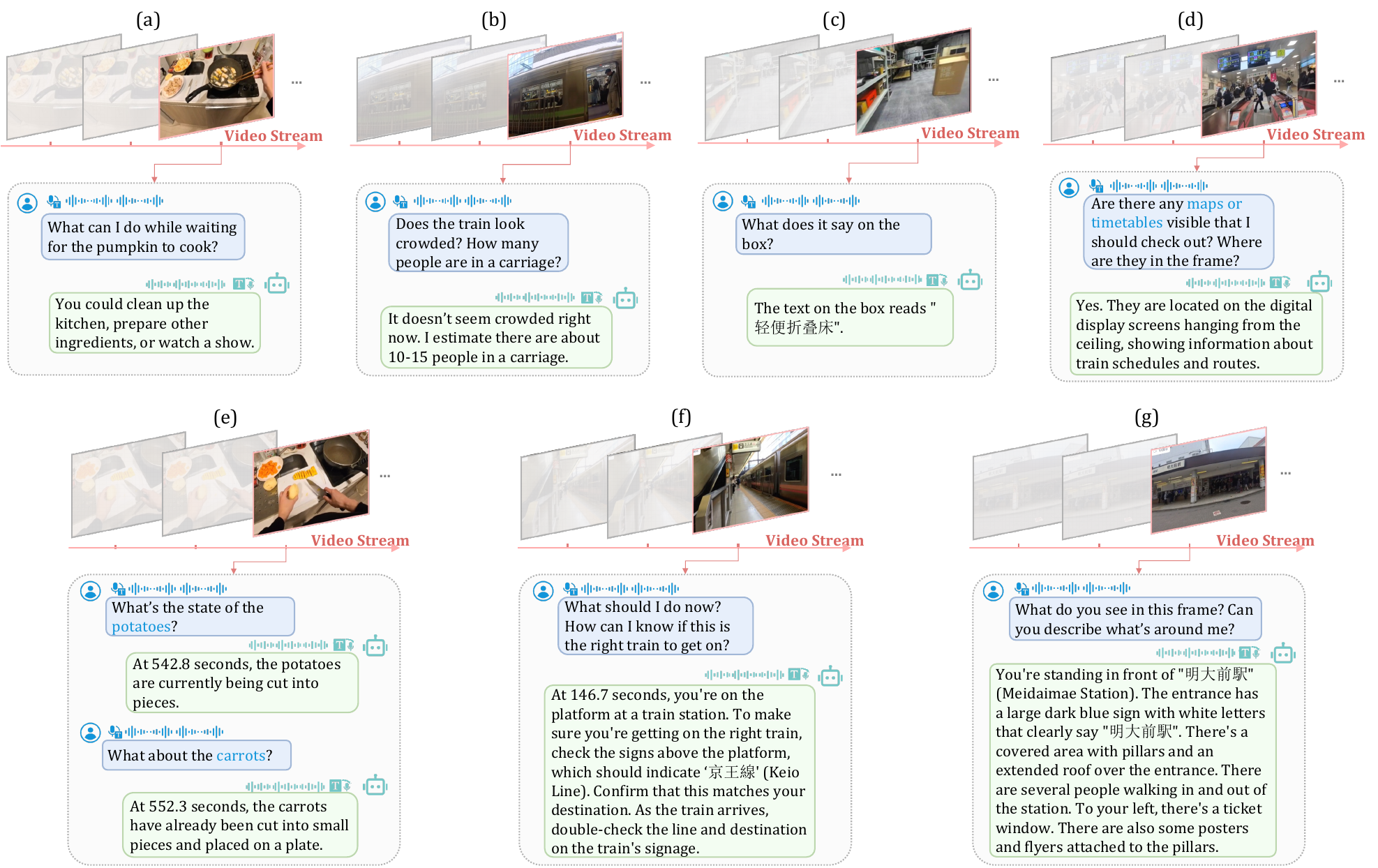}
    \caption{\textbf{Qualitative Examples of Vinci’s Chatting Functionality.} The figure illustrates seven cases (a-g) of user interactions with Vinci’s chatting functionality. For each case, the top panel shows the egocentric video captured during the interaction, while the bottom panel displays the transcribed chat log. User queries are highlighted in blue chatboxes, and Vinci’s responses are shown in green chatboxes. The interactions were conducted via audio, and the transcribed text is presented for clarity.}
    \label{fig:current}
\end{figure}

We show qualitative examples of the video, user's query and Vinci's response in Figure~\ref{fig:current}. 
In (a), Vinci demonstrates its ability to offer general knowledge assistance by suggesting productive activities such as cleaning, preparing other ingredients, or watching a show while waiting for a pumpkin to cook. Figure~\ref{fig:current}(b) highlights Vinci’s capability in crowd estimation, where it correctly assesses the number of people in a train and identifies the status as “not crowded.” Vinci also showcases strong optical character recognition (OCR) and reasoning abilities in Figure~\ref{fig:current}(c), (f) and (g). It accurately recognizes Chinese text in (c) and Japanese/English text in (f,g), integrating the extracted textual information into its reasoning process to generate useful suggestions.
The case in (d) shows Vinci can effectively interpret visual context to provide answers based on the scene, demonstrating the strength of its vision-language model. Figure~\ref{fig:current}(e) highlights Vinci’s ability to sustain multi-round dialogues, where it maintains context over long interactions ($>500s$) and provides coherent responses across multiple exchanges.
These qualitative findings align with the overall user satisfaction results in the first row of Table~\ref{tab:user}: 60\% of participants rated their experience as “satisfactory”.
Participants particularly praised Vinci's ability to accurately give answers based on the visual information such as scene and text in the scene. However, some participants noted that responses were occasionally too detailed, suggesting that response brevity could be improved for a more streamlined user experience.

\input{tabs/user_study}

The results confirm that Vinci’s chatting functionality effectively integrates vision-language understanding and real-time response generation to deliver meaningful and context-aware interactions. Its fast response speed and high user satisfaction indicate that Vinci is well-suited for real-world use as a smart assistant. As chatting encapsulates Vinci’s overall capabilities, these results validate the system's robustness and usability.

\subsubsection{Temporal grounding}
Temporal grounding is a critical functionality in Vinci, enabling the system to associate user queries with specific moments in video history. This function is required by a majority of participants in our pre-experiment user study. This capability ensures accurate responses in scenarios where context from past actions or events is required. In this evaluation, we assess Vinci’s ability to locate and interpret temporally relevant information efficiently and accurately.

Similar to the evaluation of the chatting functionality, we designed both controlled experiments and real-world environment experiments to evaluate temporal grounding.

\noindent\textbf{Controlled evaluation.$\quad$} 
In the controlled evaluation, we leveraged the same ten types of objects used in the previous experiment. Participants sequentially interacted with each object, and the order of interactions was randomly shuffled. For each trial, we queried Vinci five times with the question, “When did I interact with [object]?” and evaluated whether Vinci could successfully locate the correct temporal moment. A correct temporal grounding is defined as Vinci outputting a timestamp where the user was indeed interacting with the corresponding object. We include cases where we separately interact with the same object twice, and we count the temporal grounding to be successful only if Vinci can output all interaction instances correctly. In total, we conduct 15 trials with the duration of each resulting video ranging from 1.3 to 5.0 minutes. To analyze performance across different temporal scales, we sort these videos by their temporal length and group them into three groups: short [0 - 2.5), medium [2.5 - 3.5), and long [3.5 - 5] minutes. 

\begin{figure}
    \centering
    \includegraphics[width=\linewidth]{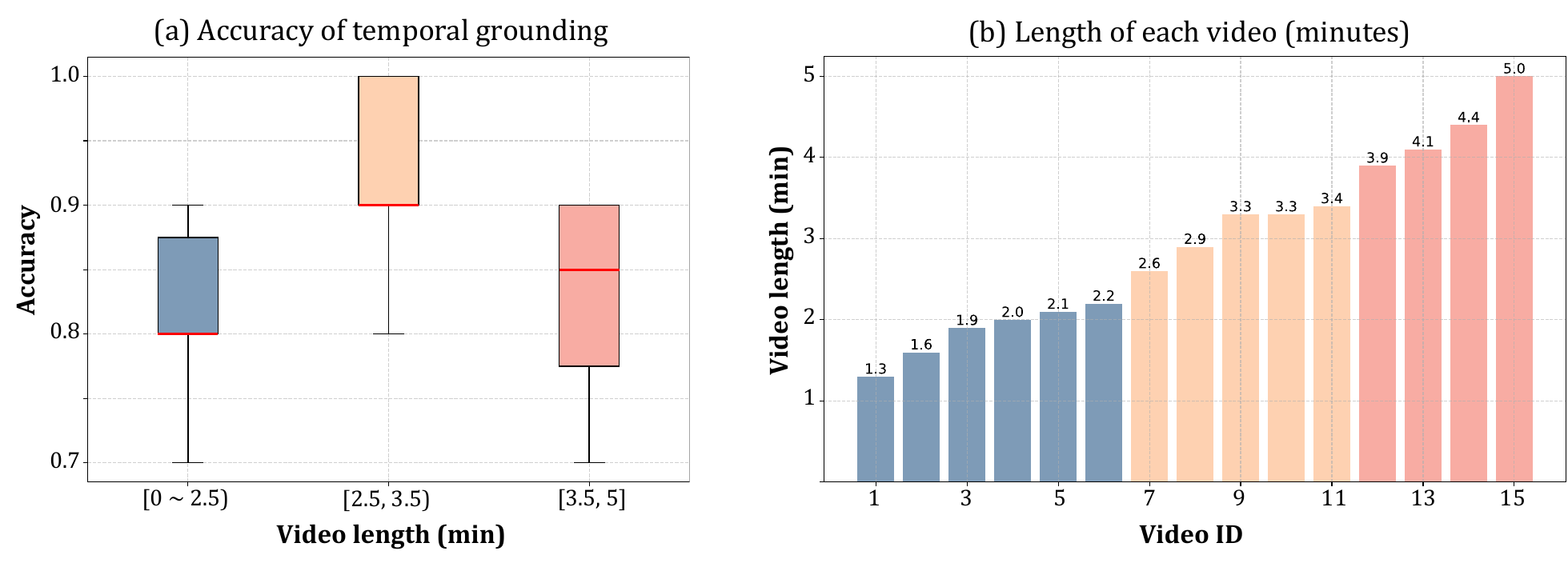}
    \caption{\textbf{Controlled Evaluation of Vinci’s Temporal Grounding Functionality.} (a) Box plot showing the accuracy of temporal grounding across videos grouped by length: short (0-2.5 minutes), medium (2.5-3.5 minutes), and long (3.5-5 minutes). The plot provides a comprehensive understanding of how video length impacts Vinci’s performance in temporally locating relevant moments. (b) Distribution of video lengths used in the evaluation, illustrating the diversity of the dataset. }
    \label{fig:grounding_controlled}
\end{figure}

\noindent\textbf{Results.$\quad$} The result of the controlled experiment for temporal grounding is shown in Figure~\ref{fig:grounding_controlled}. We can see that Vinci can maintain high accuracy across all video lengths, validating the effectiveness of our design of the memory module. The overall temporal grounding accuracy exceeded 80\%, confirming its ability to retrieve relevant historical information even in randomized, multi-instance scenarios.

\noindent\textbf{In-situ user study.$\quad$} For the real-world environment, we follow the evaluation of chatting to also consider the indoor and outdoor scenarios. In the indoor scenario, participants performed a series of tasks, such as preparing curry, arranging boxes, or making tea. The participants naturally involve in queries requiring temporal grounding such as:
“Have I added the salt? When?”,  or “When did pick up this knife?”.
In the outdoor scenario, participants navigated crowded train stations and asked temporal grounding-related queries like: “When did I enter the station?”, or “What was I doing at [specific time]?”. We evaluated accuracy similarly to the controlled experiments while also measuring response latency, response relevance, clarity, and overall user satisfaction.

\noindent\textbf{Results.$\quad$}
The quantitative results can be found in the 2nd row of Table~\ref{tab:user}. Vinci achieves over 80\% temporal grounding accuracies in both indoor and outdoor scenarios. This demonstrates the robustness of the EgoVideo-VL model and the effectiveness of the memory module. The response latency ranged from 0.6 to 0.7 seconds, which is slightly lower than in chatting tasks due to the concise nature of temporal grounding responses.

\begin{figure}
    \centering
    \includegraphics[width=\linewidth]{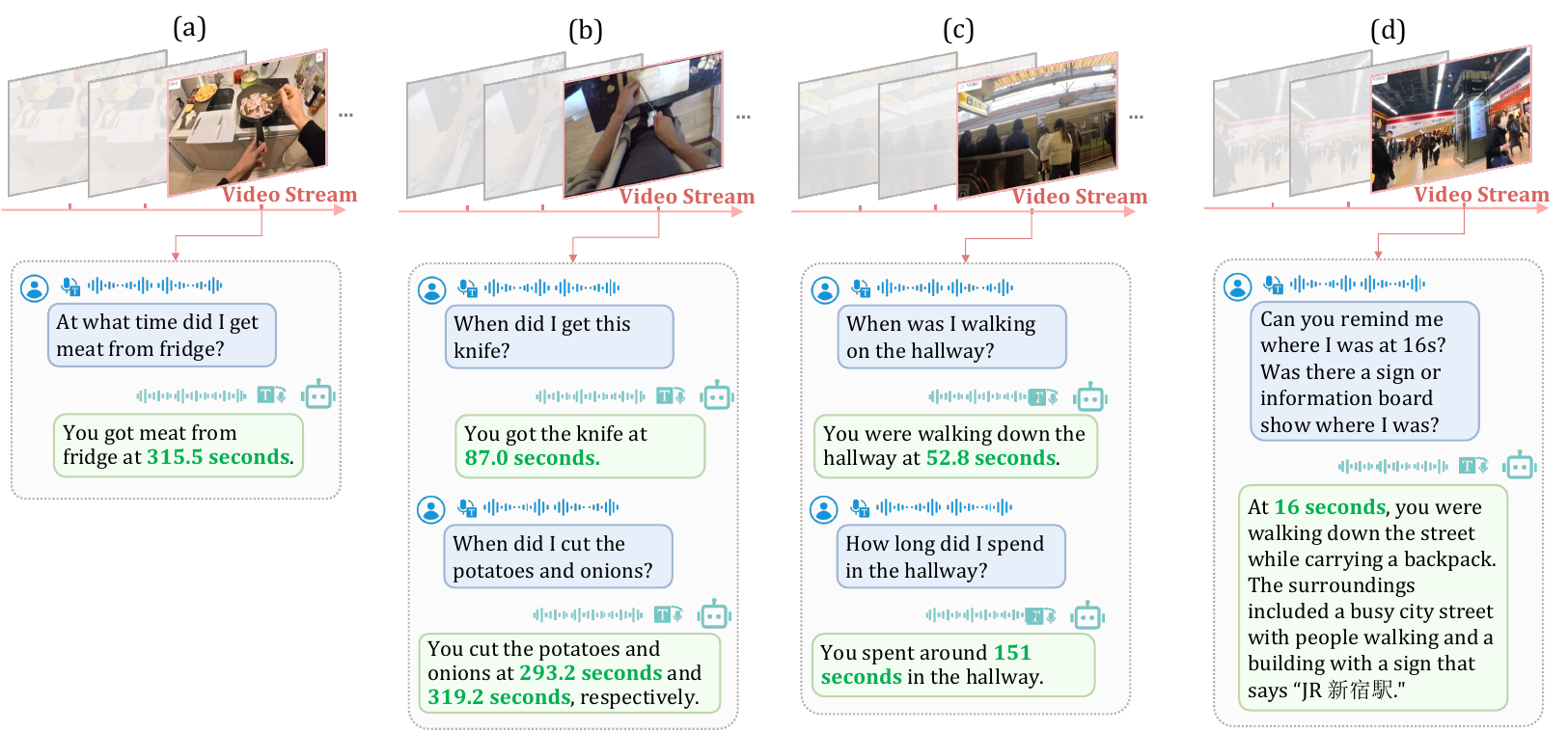}
    \caption{\textbf{Qualitative Examples of Vinci’s Temporal Grounding Functionality.} Four cases (a-d) of user interactions with Vinci’s temporal grounding functionality are illustrated. For each case, the user query is highlighted in blue. Vinci’s response identifying the corresponding temporal moment in the video is shown in green chatboxes.}
    \label{fig:ground}
\end{figure}

In Figure~\ref{fig:ground} we show quantitative examples of in-situ studies for the temporal grounding function. In (a) and (b), Vinci successfully grounds user queries in an indoor kitchen scenario. Notably, (b) demonstrates Vinci’s ability to retrieve multiple relevant moments simultaneously. In (c) and (d), outdoor examples highlight additional capabilities. In (c), Vinci estimates the duration of past events, such as the time spent walking through a hallway. In (d), Vinci helps recall specific events at a given timestamp, aligning with user expectations from our pre-experiment survey in Section 3. User feedback further supports these findings, with temporal grounding receiving high satisfaction ratings: 4.5 in indoor settings and 4.0 in outdoor environments. These results confirm Vinci’s reliability in retrieving past events efficiently, reinforcing its utility as a real-time, memory-augmented smart assistant.

\subsubsection{Summarization}
Summarization is a key functionality of Vinci, enabling the system to condense video history into a concise and coherent list of past actions. This capability is particularly useful for users who need to review or recall sequences of events. In this evaluation, we also assess Vinci’s ability to accurately and comprehensively summarize video content in both controlled setting and real-world experiments.

\noindent\textbf{Controlled evaluation.$\quad$} We evaluated Vinci’s summarization capability using the same recorded video dataset from the temporal grounding experiments, which contains interactions with ten objects: pen, pencil, scissors, cup, umbrella, toy, mouse, calculator, toothbrush, and cards. The video captures sequential interactions with these objects, with the order of interactions randomly shuffled. For each trial, we asked Vinci to “Summarize and list all the previous actions based on the history” and evaluated the quality of the generated summary.

To quantify the performance of Vinci’s summarization, we used the following metrics: 
\textbf{1. Edit distance}. We employ the Levenshtein distance, which measures the minimum number of edits (insertions, deletions, or substitutions) required to transform the generated event sequence into the ground truth event sequence. A lower value indicates a closer match to the ground truth sequence. This represents an overall accuracy of the summarization result. \textbf{2. Duplicated events}. The number of actions that are repeated in the summary, lower is better. We use this metric since we found in our experiments that Vinci sometimes outputs duplicated summaries. We did not include an ordering metric, as no significant sequencing errors were observed in our experiments.

\begin{figure}
    \centering
    \includegraphics[width=\linewidth]{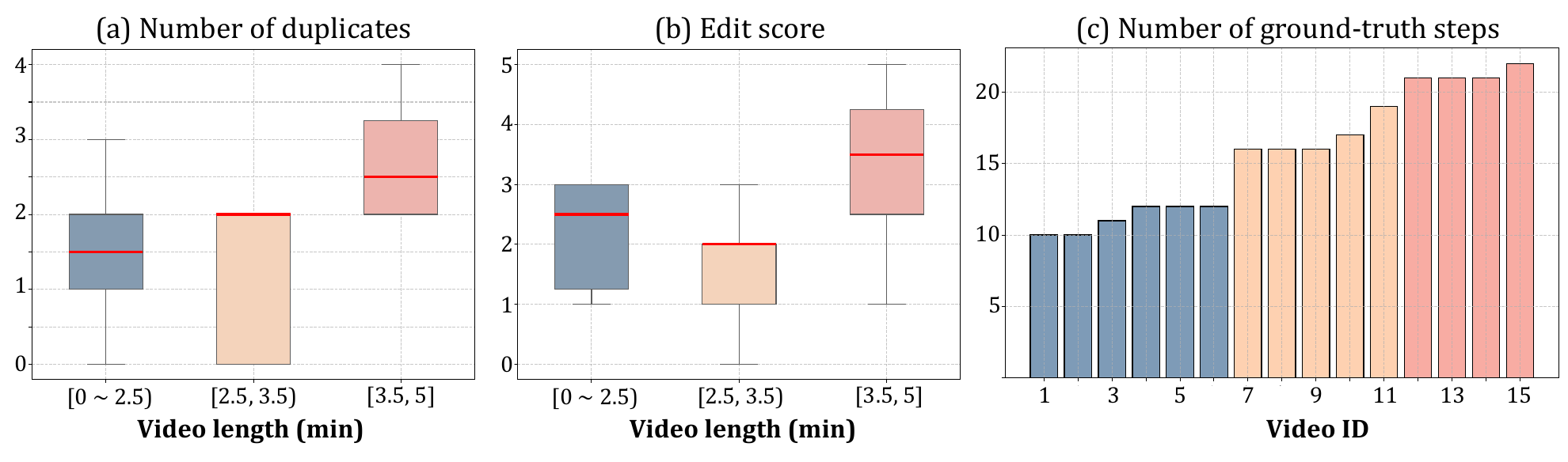}
    \caption{\textbf{Controlled Evaluation of Vinci’s Summarization Functionality.} Results from controlled experiments evaluating Vinci’s ability to summarize video content. (a) Number of duplicated events in Vinci’s summaries, (b) Edit distance scores measuring the discrepancy between Vinci’s summaries and ground truth, and (c) Number of ground truth steps in the original videos. For all subplots, videos are grouped by length into three categories: short (0-2.5 minutes), medium (2.5-3.5 minutes), and long (3.5-5 minutes) to enable a more detailed analysis of performance across varying video durations. }
    \label{fig:sum_controlled}
\end{figure}

\noindent\textbf{Results.$\quad$} The results of the controlled experiment are shown in Figure~\ref{fig:sum_controlled}. Since lower scores indicate better performance for both metrics, we can see that Vinci maintains strong summarization accuracy across different video lengths. For short and medium-length videos, the average edit distance is around 2, which is a strong result given that each summary consists of at least 10 events. However, performance slightly declines for longer videos, where Vinci occasionally misses some steps or repeats certain actions multiple times. This trend is also reflected in the duplicate count, where longer videos exhibit a higher number of repeated actions.

\noindent\textbf{In-Situ user study.$\quad$} Following the evaluations of chatting and temporal grounding, we conducted an in-situ user study to assess Vinci’s summarization performance in real-world scenarios. During each session, participants interacted with Vinci and requested a summary of their recent actions or key steps. To ensure fairness in evaluation, we limited the maximum number of summarized actions to 5. If the total number of relevant actions exceeded this limit, only the most recent 5 were considered. Accuracy was computed as the proportion of correctly summarized items relative to the total requested.

\begin{figure}
    \centering
    \includegraphics[width=\linewidth]{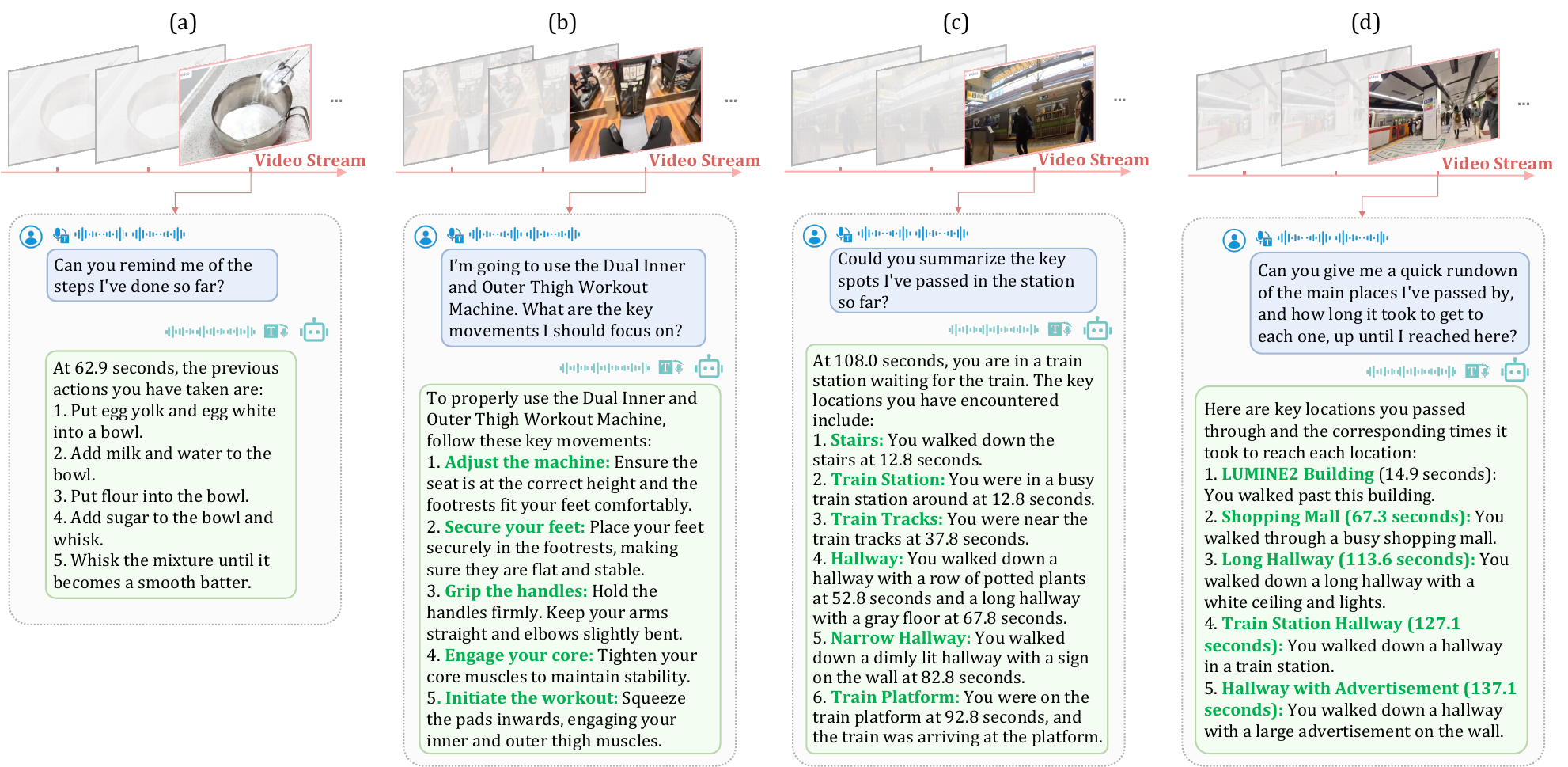}
    \caption{\textbf{Qualitative Examples of Vinci’s Summarization Functionality.} For each case, the user queries are in blue chatboxes, and Vinci's responses are shown in green chatboxes. The results underscore Vinci’s potential as a practical tool for summarizing egocentric video content in real-world scenarios.}
    \label{fig:summarize}
\end{figure}

\noindent\textbf{Results.$\quad$} As shown in the third row of Table~\ref{tab:user}, Vinci achieved over 80\% accuracy in summarization for indoor scenarios and around 70\% for outdoor environments. These results align with findings from the controlled study. For qualitative evaluation, Figure~\ref{fig:summarize} presents four example cases. In (a) and (b), Vinci demonstrates strong summarization ability in indoor environments such as a kitchen and a gym. In (c) and (d), Vinci provides outdoor summarization with corresponding timestamps, further showcasing its ability to track and structure past events. This function was highly requested by users in the pre-experiment survey, reinforcing its importance for real-world applications.

\subsubsection{Future Planning}
The future planning function is a core feature of Vinci, enabling the system to generate actionable and contextually relevant plans based on the user’s current environment and past actions. This functionality is particularly valuable in scenarios where users require step-by-step guidance to complete tasks such as cooking, organizing a workspace, or navigating an unfamiliar environment. Unlike objective tasks such as object recognition or temporal grounding, future planning is inherently open-ended, as there is no single “correct” plan for a given task. Plans often vary depending on user preferences, available resources, and contextual constraints. For this reason, we focus on in-situ user studies to evaluate Vinci’s future planning capabilities, as they provide a more realistic and meaningful assessment of the system’s utility in real-world scenarios.

\noindent\textbf{In-Situ user study.$\quad$} Following the methodology used for evaluating Vinci’s chatting functionality, we tested future planning in both indoor and outdoor scenarios. Participants interacted with Vinci naturally, requesting guidance when needed. After completing each task, they assessed the quality of the generated plans. Each plan consists of multiple steps, and accuracy was calculated as the proportion of correctly suggested steps relative to the total number of generated steps.

\begin{figure}
    \centering
    \includegraphics[width=\linewidth]{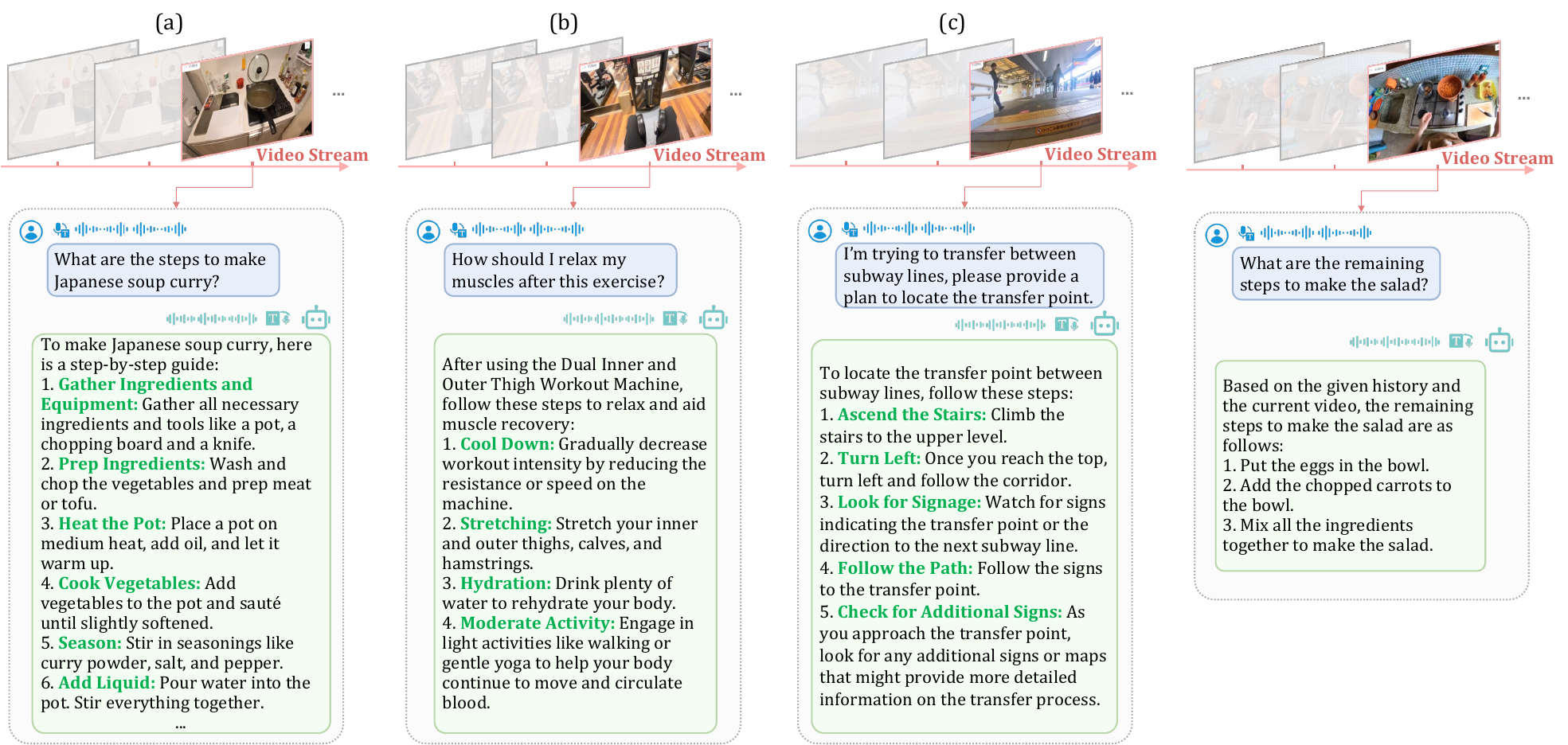}
    \caption{\textbf{Qualitative Examples of Vinci’s Future Planning Functionality.} The figure showcases three cases of Vinci’s future planning outputs across different environments: (a) indoor kitchen, (b) indoor gym, and (c) outdoor train station. For each case, the user’s query is highlighted in a blue chatbox, and Vinci’s generated plan is shown in a green chatbox. }
    \label{fig:future}
\end{figure}

\noindent\textbf{Results.$\quad$}
The quantitative results can be found in the fourth row of Table~\ref{tab:user}. User evaluations indicate that Vinci achieves an overall planning accuracy of 80.2\% in indoor scenarios and 72.7\% in outdoor scenarios.  Participants also rated the relevance and clarity of Vinci’s planning highly. Although the average response latency for this function exceeds 5 seconds, users reported that this delay was not a significant concern, as multi-step planning typically does not require immediate responses. The overall user rating for future planning was 4.3 to 4.4 on a 5-point scale, further demonstrating the system’s effectiveness.

Figure~\ref{fig:future} shows some examples of planning results when users interacted with Vinci.  In (a), Vinci provides a structured step-by-step guide for making Japanese soup curry. Since the response is lengthy (10 steps), only the first 6 steps are displayed for clarity. In (b), Vinci successfully adapts to an indoor gym scenario, assisting the user in planning an exercise routine. In (c), Vinci demonstrates spatial reasoning in an outdoor train station, advising the user to turn left after ascending the stairs to reach a transfer corridor. This ability to integrate environmental context into planning reinforces the high user satisfaction ratings and showcases Vinci’s potential as a smart assistant.

\subsubsection{Action Prediction}
The action prediction functionality enables Vinci to generate visual demonstrations of future actions based on the user’s current context. This feature is powered by the Generation module in EgoVideo-VL, which predicts and renders sequences of actions to provide intuitive, visual guidance. In this evaluation, we assess the performance of Vinci’s action prediction through controlled experiments and in-situ user studies.

\noindent\textbf{Controlled evaluation.$\quad$} To quantitatively evaluate the Generation module, we conducted experiments using the EPIC-Kitchens-100 dataset, a benchmark for egocentric action recognition and prediction. We selected five common actions: cut, open, stir, take, and turn. For each action, we randomly sampled 100 instances and evaluated Vinci’s generation performance using the following metrics: \textbf{SSIM} (Structural Similarity Index Measure), which measures the perceptual similarity between the generated video and the ground truth. \textbf{PSNR} (Peak Signal-to-Noise Ratio), which quantifies the quality of the generated video relative to the ground truth. \textbf{LPIPS} (Learned Perceptual Image Patch Similarity), which evaluates the perceptual quality of the generated video using a deep learning-based metric. And finally \textbf{FVD} (Fréchet Video Distance), which assesses the overall quality and realism of the generated video by comparing feature distributions.

\input{tabs/generation}

\noindent\textbf{Results.$\quad$} We compared Vinci’s Generation module with SEINE, a state-of-the-art image-to-video generation model. The results are summarized in Table~\ref{tab:gen}. Vinci’s Generation module outperformed SEINE in most cases across all metrics: Vinci achieved an average SSIM and PSNR of 0.381 and 13.741 compared to SEINE’s 0.374 and 13.161. For the LPIPS and FVD metrics, both of which are better when the values are lower, Vinci achieves 0.476 and 2068.266, respectively, while SEINE achieves 0.481 and 2445.974. These results demonstrate that Vinci’s Generation module is capable of producing high-quality, realistic action predictions, outperforming a state-of-the-art baseline in most scenarios of egocentric actions.

\noindent\textbf{In-Situ user study.$\quad$} To evaluate the real-world usability of the action prediction functionality, we conducted an in-situ user study in indoor environments. We do not study this functionality in outdoor environments, since the generation module of Vinci is not trained with outdoor data. Participants interacted with Vinci to receive visual demonstrations of actions such as opening a container, stirring a mixture, or cutting vegetables. We focused on two key metrics: \textbf{Latency}, \textbf{i.e.}, the time taken by Vinci to generate and display the action prediction. \textbf{User Overall Preference Score}, where participants rated their satisfaction with the action prediction functionality on a 5-point Likert scale.

\noindent\textbf{Results.$\quad$} The results can be found in the 5-th row of Table~\ref{tab:user}. The average latency for action prediction was 11.5 seconds, which participants found to be slower than expected. Participants gave an average rating of 3/5, indicating moderate satisfaction. While some users appreciated the visual demonstrations for their ability to provide intuitive guidance, others noted that the latency and occasional lack of realism reduced the functionality’s usefulness. For instance, in tasks requiring precise timing (e.g., stirring a mixture or cutting vegetables), the delay in generating predictions made the feature less practical for real-time assistance. 

\begin{figure}
    \centering
    \includegraphics[width=\linewidth]{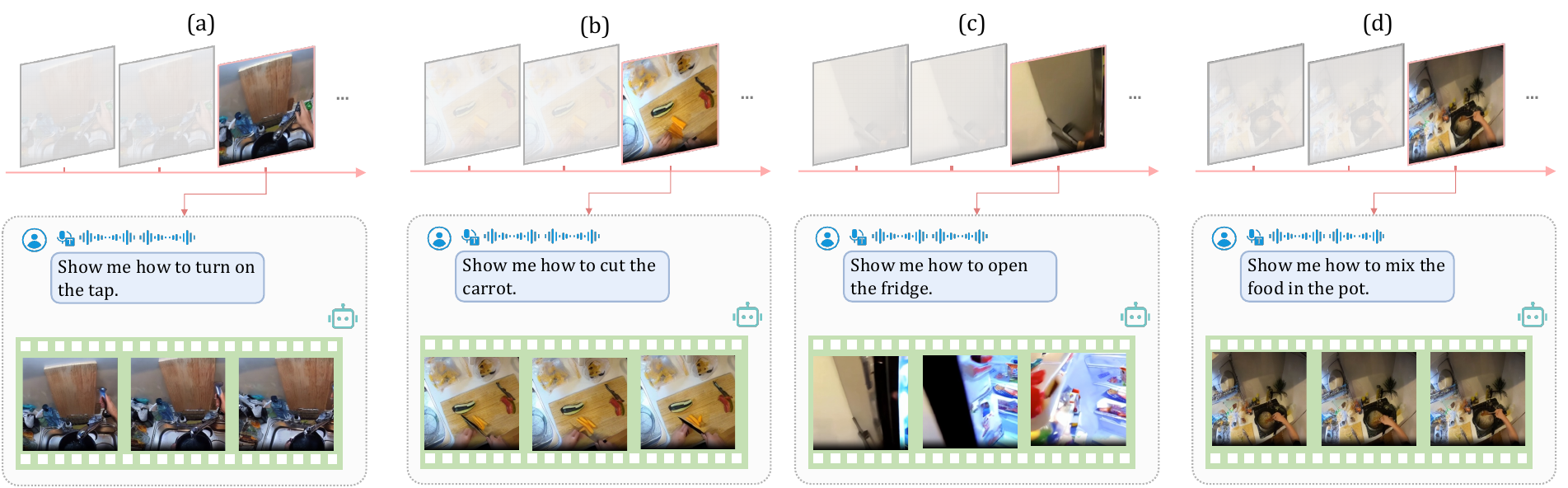}
    \caption{\textbf{Qualitative Examples of Vinci’s Action Prediction Functionality.} The figure demonstrates Vinci’s ability to generate visual demonstrations of predicted actions based on the current video frame and user queries (shown in blue chatboxes). For each case, Vinci outputs a 2-second video visualizing the predicted action, providing an intuitive how-to demo for the user.}
    \label{fig:gen}
\end{figure}

These limitations are primarily due to the computational complexity of the current Generation module, which affects real-time usability. Despite these challenges, qualitative examples of Vinci’s generation results, shown in Figure~\ref{fig:gen}, demonstrate its potential. For example, the module successfully generated realistic sequences for actions like (a) turning on a tap and (b) cutting a carrot, with accurate motion and object interactions. However, in cases like (c), the generated videos exhibited some artifacts or inconsistencies, which contributed to the perceived lack of realism. These examples highlight both the strengths and areas for improvement in Vinci’s action prediction capabilities.

It is important to note that Vinci’s modular design allows for easy integration of stronger generation modules as they become available. Future improvements in video generation technology could significantly enhance the action prediction functionality, making it more responsive and realistic for real-world applications. Despite the current limitations, the results validate the potential of Vinci’s action prediction as a valuable feature for providing visual guidance in egocentric AI systems.

\subsubsection{Video Retrieval}
The video retrieval functionality enables Vinci to bridge egocentric and third-person perspectives, allowing users to find relevant demonstration videos for tasks they are performing. This feature is powered by the Retrieval module in EgoVideo-VL, which leverages cross-view retrieval to match egocentric videos with third-person instructional videos. In this evaluation, we assess the performance of Vinci’s video retrieval through controlled experiments and in-situ user studies.

\noindent\textbf{Controlled evaluation.$\quad$}
To quantitatively evaluate the Retrieval module, we constructed a custom retrieval dataset comprising six types of actions: add (200 samples), chop (77 samples), pour (138 samples), place (187 samples), fry (62 samples), and mix (167 samples). The egocentric videos were sourced from the EPIC-Kitchens100 dataset, while the corresponding demonstration videos were collected from the HowTo100M dataset. 
\input{tabs/retrieval}

We evaluated retrieval performance using the following metrics: \textbf{Recall@1, Recall@5, Recall@10}, showing the percentage of queries for which the correct video is retrieved within the top 1, 5, or 10 results. \textbf{MeanR} (Mean Rank), which shows the average rank of the correct video in the retrieved results across all queries. A lower MeanR indicates better performance since the correct item appears earlier in the ranked list. MedianR (Median Rank), which is the median rank of the correct item across all queries. Unlike MeanR, which can be sensitive to outliers, MedianR provides a more robust estimate of ranking performance.

We compared Vinci’s Retrieval module with two strong baselines. \textbf{CLIP}~\cite{clip}: A state-of-the-art vision-language model trained on image-text pairs, using single frame input.
\textbf{CLIP-Video}~\cite{clip}: A variant of CLIP, using 4-frame input similar to our model.
%fine-tuned on video-text pairs similar to our dataset.

\noindent\textbf{Results.$\quad$} The results are summarized in Table~\ref{tab:retrieval}. Vinci’s Retrieval module significantly outperformed both baselines across all metrics. These results demonstrate that Vinci’s cross-view retrieval design is highly effective in matching egocentric videos with relevant third-person demonstrations, outperforming state-of-the-art baselines. The controlled evaluation demonstrates that Vinci’s Retrieval module is highly effective in matching egocentric videos with relevant third-person demonstrations, significantly outperforming CLIP and CLIP-Video. This success is attributed to the cross-view retrieval design, which explicitly bridges the gap between egocentric and third-person perspectives.

\noindent\textbf{In-Situ user study.$\quad$} To evaluate the real-world usability of the video retrieval functionality, we conducted an in-situ user study in an indoor kitchen scenario. Participants interacted with Vinci to retrieve demonstration videos for tasks such as chopping vegetables, pouring liquids, or mixing ingredients. Similar to the evaluation of the action prediction, we focused on two key metrics: latency and overall user satisfaction. 

\noindent\textbf{Results.$\quad$} The results can be found in the last row of Table~\ref{tab:user}. The average latency for video retrieval was 1.3 seconds, indicating fast and responsive performance. Participants gave an average rating of 3.8/5, reflecting positive feedback on the relevance and usefulness of the retrieved videos. 

\begin{figure}
    \centering
    \includegraphics[width=\linewidth]{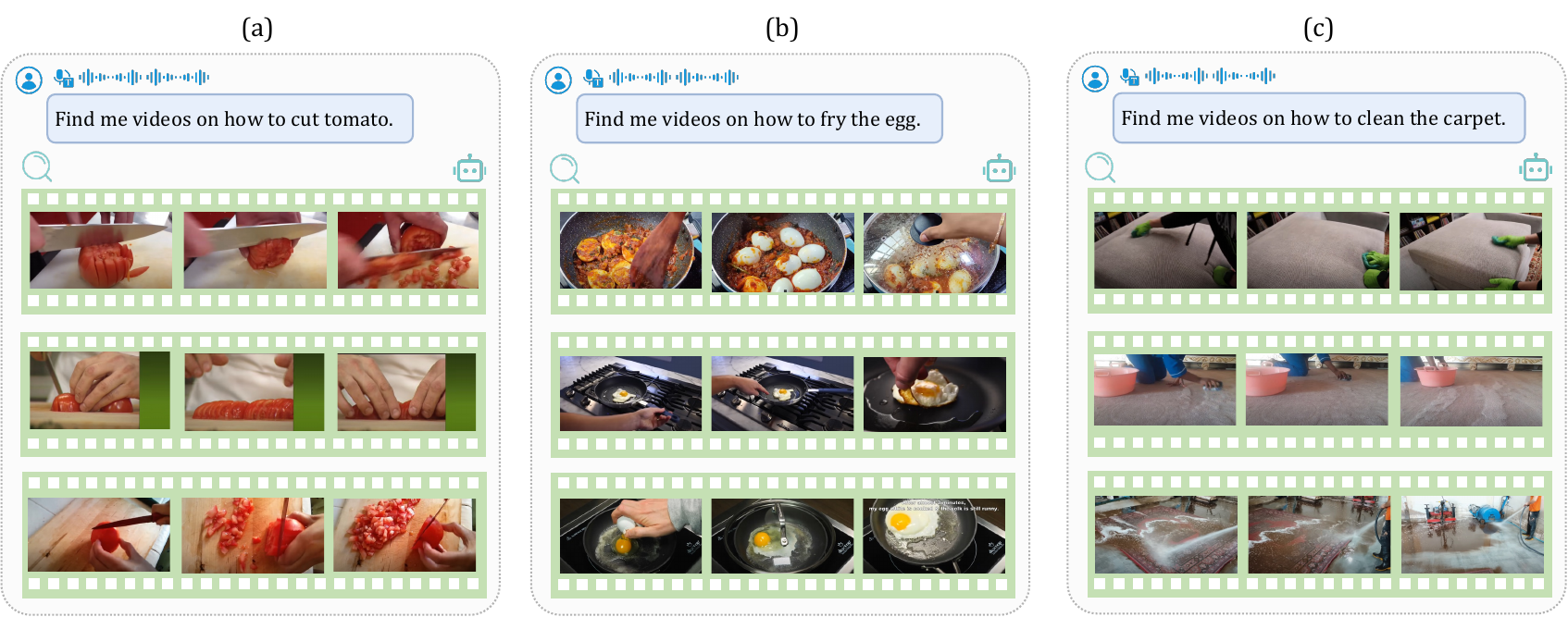}
    \caption{\textbf{Qualitative Examples of Vinci’s Video Retrieval Functionality.} The figure illustrates three cases of Vinci’s ability to retrieve relevant third-person demonstration videos based on egocentric input. For each case, Vinci successfully retrieves 3 third-person videos that align with the user’s context and query.}
    \label{fig:retrieval}
\end{figure}

Qualitative examples of Vinci’s retrieval results are shown in Figure~\ref{fig:retrieval}. The figure illustrates successful cases where Vinci accurately retrieves relevant third-person demonstration videos for egocentric actions such as cutting tomatoes, frying eggs, and cleaning the carpet. For instance, in Figure~\ref{fig:retrieval}(a), a user’s request for a video on how to cut tomatoes is correctly matched with three third-person videos demonstrating various chopping techniques. These examples highlight the effectiveness of Vinci’s cross-view retrieval design in bridging egocentric and third-person perspectives, enabling users to access diverse and relevant instructional content for their tasks.

The in-situ user study further validates the practical utility of the video retrieval functionality, with low latency and high user satisfaction scores. Participants appreciated the relevance of the retrieved videos, though some noted occasional mismatches in highly specific or nuanced tasks. These insights suggest opportunities for further refinement, such as incorporating fine-grained action recognition or user feedback to improve retrieval accuracy. Overall, the results highlight Vinci’s potential as a powerful tool for skill acquisition and task guidance in real-world scenarios.

\section{Conclusion}
In this paper, we introduced Vinci, a vision-language system designed to provide real-time, comprehensive AI assistance on portable devices. Powered by EgoVideo-VL, a novel egocentric vision-language model, Vinci integrates robust functionalities such as scene understanding, temporal grounding, video summarization, future planning, and cross-view video retrieval. Through extensive evaluations, including controlled experiments and in-situ user studies, we demonstrated Vinci’s effectiveness in delivering context-aware, actionable insights across diverse real-world scenarios.

Our experiments showcased the superior performance of EgoVideo-VL on multiple public benchmarks, highlighting its advanced vision-language reasoning and contextual understanding capabilities. User studies further validated Vinci’s practical utility, with participants praising its ability to provide intuitive guidance, retrieve relevant instructional videos, and generate actionable plans. Notably, Vinci’s hardware-agnostic design enables seamless deployment across a wide range of devices, including smartphones and wearable cameras, making it accessible to a broad user base.

Despite these successes, our evaluations also revealed areas for improvement. For instance, the action prediction functionality, while promising, faces challenges in real-time usability due to computational constraints. Similarly, the video retrieval module, though effective, could benefit from finer-grained action recognition to handle nuanced tasks. These limitations highlight exciting opportunities for future work, such as integrating more efficient generation models, enhancing cross-view retrieval with user feedback, and expanding Vinci’s capabilities to support a wider range of environments and tasks.

In conclusion, Vinci represents a significant step forward in the development of portable, real-time egocentric AI systems. By combining state-of-the-art vision-language modeling with practical, user-centric design, Vinci empowers users with contextual and actionable insights, paving the way for a new generation of smart assistants. We hope that Vinci will inspire further research in egocentric AI and contribute to the creation of intelligent systems that seamlessly integrate into our daily lives.

\bibliographystyle{ACM-Reference-Format}
\bibliography{sample-base}
\end{document}

%% file: tabs/user_study.tex
\renewcommand{\arraystretch}{1.2}
\begin{table}[t!]

\caption{\textbf{Results of the In-Situ User Study on Vinci’s Functionalities.} The table summarizes user study evaluations for Vinci’s six core functionalities: Chatting, Temporal Grounding, Summarization, Future Planning, Action Prediction, and Video Retrieval. For the first four functionalities, we report metrics including accuracy, latency, relevance (rated 1-5), clarity (rated 1-5), and overall satisfaction score (rated 1-5), evaluated in both indoor and outdoor scenarios. For Action Prediction and Video Retrieval, we focus on latency and overall satisfaction and assessed Vinci only in indoor scenarios.} 
\label{tab:user}
\centering
\resizebox{\linewidth}{!}{
% \vspace{-0.1in}
    \begin{tabular}{l|cc|cc|cc|cc|cc}
 \Xhline{1.0pt}
    \rowcolor{mygray}
     & \multicolumn{2}{c|}{\bf Accuracy (\%)} & \multicolumn{2}{c|}{\bf Latency (s)} &  \multicolumn{2}{c|}{\bf Relevance (1-5)} & \multicolumn{2}{c|}{\bf Clarity (1-5)} & \multicolumn{2}{c}{\bf Overall (1-5)}  \\
     \rowcolor{mygray} \multirow{-2}{*}{\bf Task} &\textit{Indoor} & \textit{Outdoor}  &\textit{Indoor} & \textit{Outdoor} &\textit{Indoor} & \textit{Outdoor} &\textit{Indoor} & \textit{Outdoor} &\textit{Indoor} & \textit{Outdoor} \\
\Xhline{0.7pt}
Chat & 91.0 & 84.0 & 0.8 & 0.6 & 5.0 & 5.0 & 4.5 & 4.1 & 4.8 & 4.4 \\
Temporal grounding & 87.0 & 82.0 & 0.6 & 0.7 & 4.8 & 4.2 & 4.9 & 4.5 & 4.5 & 4.0 \\
Summarization & 80.2 & 72.7 & 8.0 & 9.4 & 4.2 & 3.7 & 3.5 & 3.3 & 3.5 & 3.2 \\
Planning & 84.1 & 76.3 & 5.4 & 6.6 & 4.8 & 4.4 & 4.6 & 4.3 & 4.4 & 4.3 \\
Action prediction & - & - & 11.5 & - & - & - & - & - & 3.0 & - \\
Video retrieval & - & - & 1.3 & - & - & - & - & - & 3.8 & - \\
 \Xhline{1.0pt}
    \end{tabular}
}
\end{table}

%% file: tabs/generation.tex
% \begin{table}[t!]
% \caption{Retrieval exps} 
% \label{tab:retrieval}
% \centering
% \resizebox{\linewidth}{!}{
% \begin{tabular}{lccccccc}
%  \midrule
%      & \multicolumn{3}{c}{\bf Video$\rightarrow$Text} & \multicolumn{3}{c}{\bf  Text$\rightarrow$Video} & Average  \\
%      \cmidrule(r){2-4} \cmidrule(r){5-7} \cmidrule(r){8-8}
%      \multirow{-2}{*}{\bf Method} & R@1 & R@5 & R@10 & R@1 & R@5 & R@10 & Avg \\
% %\Xhline{0.7pt}
% \midrule

% xxx
%  \midrule
%     \end{tabular}
% }
% \end{table}

\begin{table}
\caption{\textbf{Quantitative Evaluation of Vinci’s Action Prediction Functionality.}
The table compares Vinci’s action prediction performance with SEINE~\cite{seine}, a state-of-the-art image-to-video generation model, across five actions: cut, open, stir, take, and turn. Performance is evaluated using four metrics: SSIM, PSNR, LPIPS, and FVD. $\uparrow$ indicates higher values are better and $\downarrow$ indicates lower values are better.} 
\label{tab:gen}
\centering
\resizebox{0.6\linewidth}{!}{
% \vspace{-0.1in}
    \begin{tabular}{lc|cccc}
 \Xhline{1.0pt}
     \rowcolor{mygray} \bf Model & \bf Action  & \textit{SSIM$\uparrow$} & \textit{PSNR$\uparrow$} & \textit{LPIPS$\downarrow$} & \textit{FVD$\downarrow$} \\
\Xhline{0.7pt}
SEINE~\cite{seine}  & \multirow{2}{*}{cut} & 0.378 & 13.540 & 0.455 & 2749.529 \\
\bf Vinci (EgoVideo-VL) & & 0.367 & 14.370 & 0.441 & 2159.011 \\
\Xhline{0.7pt}
SEINE~\cite{seine}  & \multirow{2}{*}{open} & 0.364 & 12.150 & 0.561 & 1995.473 \\
\bf Vinci (EgoVideo-VL) & & 0.369 & 12.442 & 0.550 & 1730.397 \\
\Xhline{0.7pt}
SEINE~\cite{seine}  & \multirow{2}{*}{stir} & 0.412 & 14.844 & 0.384 & 2717.788 \\
\bf Vinci (EgoVideo-VL) & & 0.444 & 15.043 & 0.363 & 2807.843 \\
\Xhline{0.7pt}
SEINE~\cite{seine}  & \multirow{2}{*}{take} & 0.355 & 12.719 & 0.524 & 2329.876 \\
\bf Vinci (EgoVideo-VL) & & 0.358 & 13.582 & 0.519 & 2071.345 \\
\Xhline{0.7pt}
SEINE~\cite{seine}  & \multirow{2}{*}{turn} & 0.362 & 12.550 & 0.479 & 2437.206 \\
\bf Vinci (EgoVideo-VL) & & 0.368 & 13.270 & 0.509 & 1572.736 \\
\Xhline{0.7pt}
SEINE~\cite{seine}  & \multirow{2}{*}{average} & 0.374 &	13.161	& 0.481	&2445.974\\
\bf Vinci (EgoVideo-VL) & & \textbf{0.381}	&\textbf{13.741}	&\textbf{0.476}&	\textbf{2068.266}\\
\Xhline{1.0pt}
\end{tabular}
}
\end{table}

%% file: tabs/retrieval.tex
% \begin{table}[t!]
% \caption{Retrieval exps} 
% \label{tab:retrieval}
% \centering
% \resizebox{\linewidth}{!}{
% \begin{tabular}{lccccccc}
%  \midrule
%      & \multicolumn{3}{c}{\bf Video$\rightarrow$Text} & \multicolumn{3}{c}{\bf  Text$\rightarrow$Video} & Average  \\
%      \cmidrule(r){2-4} \cmidrule(r){5-7} \cmidrule(r){8-8}
%      \multirow{-2}{*}{\bf Method} & R@1 & R@5 & R@10 & R@1 & R@5 & R@10 & Avg \\
% %\Xhline{0.7pt}
% \midrule

% xxx
%  \midrule
%     \end{tabular}
% }
% \end{table}

\begin{table}
\caption{\textbf{Quantitative Evaluation of Vinci’s Video Retrieval Functionality.} The table compares Vinci’s retrieval performance with CLIP and CLIP-Video across six actions: add, chop, fry, mix, place, and pour. Both video-to-text retrieval and text-to-video retrieval are evaluated to comprehensively assess Vinci’s cross-view retrieval capabilities. Performance is measured using Recall@1, Recall@5, Recall@10, MeanR, and MedianR. $\uparrow$ indicates higher values are better and $\downarrow$ indicates lower values are better. The results highlight the effectiveness of Vinci’s cross-view retrieval design in bridging egocentric and third-person perspectives.} 
\label{tab:retrieval}
\centering
\resizebox{\linewidth}{!}{
% \vspace{-0.1in}
    \begin{tabular}{lc|ccccc|ccccc}
 \Xhline{1.0pt}
    \rowcolor{mygray}
     & & \multicolumn{5}{c|}{\bf Video$\rightarrow$Text} &  \multicolumn{5}{c}{\bf Text$\rightarrow$Video} \\
     \rowcolor{mygray} \multirow{-2}{*}{\bf Model} & \multirow{-2}{*}{\bf Action}  & \textit{R@1$\uparrow$} & \textit{R@5$\uparrow$} & \textit{R@10$\uparrow$} & \textit{MeanR$\downarrow$} & \textit{MediumR$\downarrow$}  & \textit{R@1$\uparrow$} & \textit{R@5$\uparrow$} & \textit{R@10$\uparrow$} & \textit{MeanR$\downarrow$} & \textit{MediumR$\downarrow$} \\
\Xhline{0.7pt}
CLIP~\cite{clip}  & \multirow{3}{*}{add} & 14.4 & 31.0 & 41.5 & 33.9 & 15.0 & 12.0 & 31.6 & 41.8 & 39.7 & 15.5 \\
CLIP-Video~\cite{clip}  & & 15.0 & 34.0 & 45.3 & 33.3 & 13.0 & 14.4 & 34.7 & 43.5 & 37.3 & 15.0 \\
\bf Vinci & & 60.0 & 90.5 & 96.0 & 3.6 & 1.0 & 43.0 & 73.0 & 85.0 & 5.8 & 2.0 \\
\Xhline{0.7pt}
CLIP~\cite{clip}  & \multirow{3}{*}{chop} & 27.4 & 57.1 & 66.7 & 11.6 & 3.5 & 33.8 & 55.0 & 70.0 & 13.3 & 2.5 \\
CLIP-Video~\cite{clip}  & & 32.9 & 64.6 & 74.4 & 9.8 & 3.0 & 26.9 & 62.8 & 78.2 & 9.3 & 4.0 \\
\bf Vinci & & 54.4 & 88.6 & 94.9 & 3.7 & 1.0 & 35.1 & 71.4 & 83.1 & 6.6 & 2.0 \\
\Xhline{0.7pt}
CLIP~\cite{clip}  & \multirow{3}{*}{fry} & 23.3 & 52.1 & 58.9 & 11.5 & 5.0 & 25.0 & 56.3 &  68.8 & 11.7 & 5.0 \\
CLIP-Video~\cite{clip} & & 28.8 & 50.7 & 63.0 & 9.5 & 5.0 & 31.3 & 61.2 & 70.1 & 9.6 & 4.0 \\
\bf Vinci & & 44.1 & 80.9 & 94.1 & 3.4 & 2.0 & 33.9 & 69.4 & 80.6 & 7.3 & 2.5 \\
\Xhline{0.7pt}
CLIP~\cite{clip}  & \multirow{3}{*}{mix} & 16.6 & 30.2 & 42.2 & 29.4 & 14.0 & 16.7 & 33.9 & 50.0 & 32.9 & 10.5 \\
CLIP-Video~\cite{clip}  & & 14.4 & 34.3 & 40.7 & 25.6 & 14.0 & 22.4 & 42.2 & 52.1 & 28.1 & 9.5 \\
\bf Vinci & & 48.8 & 79.3 & 87.3 & 6.4 & 2.0 & 39.5 & 64.6 & 76.6 & 10.7 & 3.0 \\
\Xhline{0.7pt}
CLIP~\cite{clip}  & \multirow{3}{*}{place} & 23.5 & 56.4 & 67.9 & 15.8 & 5.0 & 26.1 & 51.3 & 70.4 & 17.2 & 5.0 \\
CLIP-Video~\cite{clip}  & & 30.7 & 61.0 & 79.0 & 11.3 & 3.0 & 29.2 & 62.9 & 75.2 & 12.1 & 3.0 \\
\bf Vinci & & 47.1 & 80.6 & 87.4 & 5.1 & 2.0 & 41.7 & 73.7 & 85.6 & 6.7 & 2.0 \\
\Xhline{0.7pt}
CLIP~\cite{clip}  & \multirow{3}{*}{pour} & 19.2 & 37.2 & 53.2 & 22.3 & 9.5 & 14.9 & 37.7 & 49.4 & 25.1 & 11.0 \\
CLIP-Video~\cite{clip}  & & 23.2 & 47.1 & 61.3 & 19.1 & 6.0 & 18.4 & 43.4 & 53.9 & 22.2 & 8.0 \\
\bf Vinci & & 47.9 & 77.8 & 88.9 & 4.5 & 2.0 & 37.0 & 68.8 & 81.8 & 8.7 & 2.5 \\
\Xhline{0.7pt}
CLIP~\cite{clip}  & \multirow{3}{*}{average} & 20.7 & 44.0 & 55.1 & 20.7 & 8.7 & 21.4 & 44.3 & 58.4 & 23.3 & 8.2 \\
CLIP-Video~\cite{clip} & & 24.1 & 48.6 & 60.6 & 18.1 & 7.3 & 23.7 & 51.2 & 62.1 & 19.7 & 7.2 \\
\bf Vinci & & \textbf{50.3} & \textbf{82.9} & \textbf{91.4} & \textbf{4.4} & \textbf{1.6} & \textbf{38.3} & \textbf{70.1} & \textbf{82.1} & \textbf{7.6} & \textbf{2.3}  \\
\Xhline{1.0pt}
    \end{tabular}
}
\end{table}